\documentclass[twoside,11pt]{article}
\usepackage{jair, theapa, rawfonts}

\jairheading{64}{2019}{X-XX}{03/19}{10/18}
\ShortHeadings{Pitfalls and Best Practices for Algorithm Configuration}{Eggensperger, Lindauer, \& Hutter}
\firstpageno{1}

\usepackage[utf8]{inputenc}
\usepackage{amsmath,amssymb,amsfonts,dsfont}
\usepackage{url}
\usepackage[svgnames]{xcolor}
\usepackage{booktabs}
\usepackage{multirow}
\usepackage{tabularx}
\usepackage{subcaption}
\newcommand{\denselist}{\itemsep -1.5pt\partopsep -20pt}
\usepackage{listings}

\usepackage{tikz}
\usetikzlibrary{shapes.geometric}
\usetikzlibrary{positioning,shapes,shadows,arrows,calc}
\usetikzlibrary{intersections}
\pgfdeclarelayer{background}
\pgfdeclarelayer{foreground}
\pgfsetlayers{background,main,foreground}

\newcommand{\insts}[0]{\Pi}
\newcommand{\inst}[0]{\pi}

\newcommand{\pcs}[0]{\vec{\Theta}}
\newcommand{\conf}[0]{\vec{\theta}}
\newcommand{\cutoff}[0]{\kappa}

\renewcommand{\vec}[1]{\mathbf{#1}}
\newcommand{\algo}[0]{\mathcal{A}}

\newcommand{\perf}[0]{\mathbb{R}}

\newcommand{\aclib}{\textit{AClib}} 
\newcommand{\smac}{\textit{SMAC}}
\newcommand{\dsmac}{\textit{dSMAC}}
\newcommand{\roar}{\textit{ROAR}} 
\newcommand{\paramils}{\textit{ParamILS}}

\newcommand{\gga}{\textit{GGA}} 
\newcommand{\irace}{\textit{irace}}

\newcommand{\glucose}{\textit{Glucose}}
\newcommand{\clasp}{\textit{Clasp}}
\newcommand{\lingeling}{\textit{Lingeling}}
\newcommand{\minisathack}{\mbox{\textit{Minisat-HACK-999ED}}}

\newcommand{\circuitfuzz}{\textit{Circuit Fuzz}}
\newcommand{\saps}{\textit{Saps}}
\newcommand{\cryptominisat}{\textit{Cryptominisat}}

\newcommand{\minisat}[0]{\textit{MiniSAT}}
\newcommand{\wrapper}[0]{\texttt{GenericWrapper4AC}}
\newcommand{\runsolver}[0]{\texttt{runsolver}}
\newcommand{\runtime}[0]{runtime}

\DeclareMathOperator*{\argmin}{arg\,min}

\begin{document}

\title{Pitfalls and Best Practices in Algorithm Configuration}

\author{\name Katharina Eggensperger \email eggenspk@cs.uni-freiburg.de\\
       \name Marius Lindauer \email lindauer@cs.uni-freiburg.de\\
       \name Frank Hutter \email  fh@cs.uni-freiburg.de\\
       \addr Institut für Informatik, Albert-Ludwigs-Universität Freiburg,\\
             Georges-Köhler-Allee 74, 79110 Freiburg, Germany}

% For research notes, remove the comment character in the line below.
%\researchnote

\maketitle

\begin{abstract}
Good parameter settings are crucial to achieve high performance in many areas of artificial intelligence (AI), such as propositional satisfiability solving, AI planning, scheduling, and machine learning (in particular deep learning). Automated algorithm configuration methods have recently received much attention in the AI community 
since they replace tedious, irreproducible and error-prone manual parameter tuning and can lead to new state-of-the-art performance. 
However, practical applications of algorithm configuration are prone to several (often subtle) pitfalls in the experimental design that can render the procedure ineffective. We identify several common issues and propose best practices for avoiding them. As one possibility for automatically handling as many of these as possible, we also propose a tool called \texttt{GenericWrapper4AC}.
\end{abstract}

%%%%%%%%%%%%%%%%%%%%%%%%%%%%%%%%%%%%%%%%%%%%%%%%%%%%%%%%%%%%%%%%%%%%
\section{Introduction}
\label{sec:intro}
%%%%%%%%%%%%%%%%%%%%%%%%%%%%%%%%%%%%%%%%%%%%%%%%%%%%%%%%%%%%%%%%%%%%

To obtain peak performance of an algorithm, it is often necessary to tune its parameters.
The AI community has recently developed automated methods for the resulting \emph{algorithm configuration (AC)} problem to replace tedious, irreproducible and error-prone manual parameter tuning. Some example applications, for which automated AC procedures led to new state-of-the-art performance, include satisfiability solving~(\citeR{hutter-fmcad07a}; \shortciteR{hutter-aij17a}), 
maximum satisfiability~\cite{ansotegui-aij16},
scheduling~\cite{chiarandini-patat08a},
mixed integer programming~\cite{hutter-cpaior10a,lopez-ejor14a},
evolutionary algorithms~\cite{bezerra-tec16a},
answer set solving~\shortcite{gebser-lpnmr11a},
AI planning~\cite{vallati-socs13a}
and
machine learning~\cite{thornton-kdd13a,feurer-aaai15a}.

Although the usability of AC systems improved over the years \cite<e.g., \textit{SpySMAC}, >{falkner-sat15a},
we still often observe fundamental issues in the design and execution of experiments with algorithm configuration methods by both experts and new users.
The goals of this work are therefore to:
\begin{itemize}
\denselist
	\item highlight the many pitfalls we have encountered in AC experiments (run by ourselves and others);
	\item present best practices to avoid most of these pitfalls; and  
	\item propose a unified interface between an AC system and the algorithm it optimizes (the so-called target algorithm) that directly implements best practices
	related to properly measuring the target algorithm's performance with different parameter settings. 
\end{itemize}

Providing recommendations and best practices on how to empirically evaluate algorithms and avoid pitfalls is a topic of interest cutting across all of artificial intelligence, including, e.g., evolutionary optimization~\cite{weise-jcst12a}, algorithms for NP-complete problems~\shortcite{gent-report97a}, and reinforcement learning~\shortcite{henderson-arxiv13a} to mention only a few.
Running and comparing implementations of algorithms is the most commonly used approach to understand the behaviour of the underlying method~\cite{mcgeoch-phd87}.
There is a rich literature on how to best conduct such empirical studies~\cite{hooker-jh95a,gent-report97a,howe-jair02a,mcggeoch-hgo02a,mcgeoch-book12a}, and for some journals abiding by such guidelines is even mandatory in order to publish research~\cite{dorigo-sw16,laguna-jh}.
Research in AC depends even more on proper empirical methodology than the rest of artificial intelligence, since AC systems need to \emph{automatically} evaluate the empirical performance of different algorithm variants in their inner loop in order to find configurations with better performance. Nevertheless, many of the underlying characteristics of empirical evaluations still remain the same as for other domains, and our guidelines thus share many characteristics with existing guidelines and extend them to the setting faced in AC. 

The structure of this work is as follows.
First, we provide a brief overview of AC, including some guidelines for new users, such as why and when to use AC, and how to set up effective AC experiments (Section~\ref{sec:ac}).
Afterwards, we describe common pitfalls in using AC systems and recommendations on how to avoid them. We first discuss pitfalls concerning the interface between AC systems and target algorithms (Section~\ref{sec:pitfalls_wrapper}), followed by pitfalls regarding over-tuning (Section~\ref{sec:pitfalls_overtuning}).
Throughout, we illustrate pitfalls by AC experiments on propositional satisfiability solvers~\cite{biere-satbook} 
as a prototypical AC example, but insights directly transfer to other AC problems.\footnote{
For these pitfalls, we do not distinguish between decision and optimization problems as the application domain of AC. Although configurators usually take into account whether and how the metric to be optimized relates to runtime,
% and hence behave differently, 
all presented pitfalls can happen in both types of application domain.}
From our own experiences, we provide further general recommendations for effective configuration in Section~\ref{sec:further}.
We end by presenting a package to provide an interface between AC systems and target algorithms that aims to improve the reliability, reproducibility and robustness of AC experiments (Section~\ref{sec:gen_wrapper}).

%%%%%%%%%%%%%%%%%%%%%%%%%%%%%%%%%%%%%%%%%%%%%%%%%%%%%%%%%%%%%%%%%%%%
\section{Background: Algorithm Configuration}
\label{sec:ac}
%%%%%%%%%%%%%%%%%%%%%%%%%%%%%%%%%%%%%%%%%%%%%%%%%%%%%%%%%%%%%%%%%%%%

The algorithm configuration problem can be briefly described as follows:
given an algorithm $\algo$ to be optimized (the so-called \emph{target algorithm}) with parameter configuration space $\pcs$,
a set of instances $\insts$, and a cost metric $c: \pcs \times \insts \to \perf$, find a configuration $\conf^* \in \pcs$ that minimizes the cost metric $c$ across the instances in $\insts$:

\begin{equation}
\label{eq:ACsimple}
\conf^* \in \argmin_{\conf \in \pcs}\sum_{\inst \in \insts} c(\conf,\inst).
\end{equation}

A concrete example for this algorithm configuration problem would be to find a parameter setting $\conf \in \pcs$ of a solver $\algo$ for the propositional satisfiability problem (SAT) (\citeR<such as \textit{glucose},>{audemard-ijcai09a}\citeR< or \textit{lingeling},>{biere-tech13a}) 
on a set of CNF instances~$\insts$ (e.g., SAT-encoded hardware or software verification instances) 
that minimizes $\algo$'s average \runtime{}~$c$.
Another example would be to find a hyperparameter setting for a machine learning algorithm that minimizes its error $c$ on a given dataset~\cite{snoek-nips12a,feurer-aaai15a}; in this latter example, $c$ would be 
validation error, either measured via $k$-fold inner cross-validation (giving rise to $k$ instances for algorithm configuration) or a single validation set (in which case there is just a single instance for algorithm configuration).

\begin{figure}
\scalebox{0.9}{
\tikzstyle{activity}=[rectangle, draw=black, rounded corners, text centered, text width=6em, fill=white, drop shadow]
\tikzstyle{data}=[rectangle, draw=black, text centered, fill=black!10, text width=8em, drop shadow]
\tikzstyle{myarrow}=[->, thick]
\begin{tikzpicture}[node distance=10em]

	\node (PCS) [data] {Configuration Space $\pcs$ of $\algo$};
	\node (AC) [activity, below of=PCS, node distance=4em] {Algorithm Configurator};
	\node (TA) [activity, right of=AC, node distance=21em] {Target\\ Algorithm $\algo$};
	\node (Instances) [data, right of=TA, node distance=12em] {Instance $\pi$};
	
	\draw[myarrow] (PCS) -- (AC);
	\draw[myarrow] (AC.east) -- node(Call) [above] {Call $\algo$ with $\conf\in\pcs$ on $\inst \in \insts$}  ($(TA.west)+(-0.5,0.0)$);
	%\node [below of=Call, node distance=1.7em] {with cutoff time $\kappa$};
	\node [below of=Call, node distance=1.3em] {\footnotesize{with cutoff time $\cutoff$}};
	\node [below of=Call, node distance=2.1em] {\footnotesize{and other limits}};
	
	\draw[myarrow] (TA) -- node[above] {Solves}  (Instances);
	
	\draw[myarrow] ($(TA.south)+(-0.0,-0.4)$) |- ++(0.0,-0.2) node[below,xshift=-7em] {Return Cost $c(\conf, \inst)$} -| (AC.south);
	
	\begin{pgfonlayer}{background}

        % Configuration Process
    	\path (TA -| TA.west)+(-0.5,0.8) node (resUL) {};
    	\path (Instances.east |- Instances.south)+(0.5,-0.6) node(resBR) {};
    	\path [rounded corners, draw=black!50] (resUL) rectangle (resBR);
		\path (TA.east |- TA.south)+(4.6,-0.1) node [text=black!80] {Wrapper};

    \end{pgfonlayer}
	
\end{tikzpicture}
}
\caption{Workflow of Algorithm Configuration}
\label{fig:ac}
\end{figure}
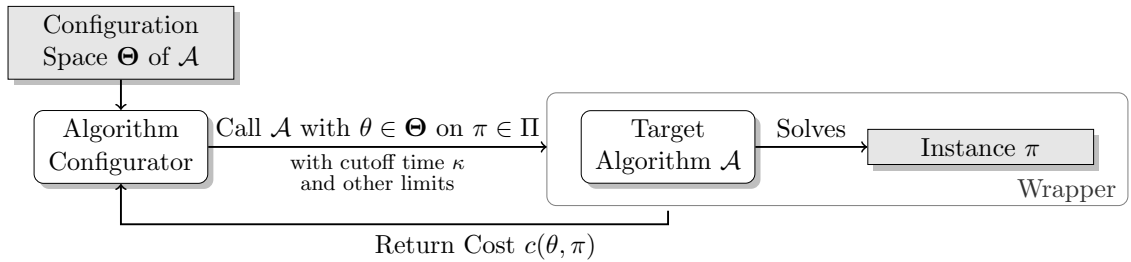

The general workflow of a sequential algorithm configuration procedure (short: \emph{configurator}) is shown in Figure~\ref{fig:ac}.
In each step, the configurator picks a configuration $\conf \in \pcs$
and an instance $\inst \in \insts$, triggers a run of algorithm $\algo$ with configuration $\conf$ on instance $\inst$ with a maximal \runtime{} cutoff $\cutoff$ 
(and other resource limitations that apply, such as a memory limit), 
and measures the resulting cost $c(\conf,\inst)$. As detailed in Section \ref{sec:gen_wrapper}, this step is usually mediated by a target-algorithm specific \emph{wrapper}.
The configurator uses this collected data about the target algorithm's performance to find a well-performing configuration, typically operating as an anytime procedure until its configuration budget is exhausted (e.g., a maximal number of target algorithm calls or a time budget)\footnote{Alternatively, the termination criterion could be defined as stopping when no (or only little) further improvement is expected. Although this is a common choice for some other anytime algorithms, such as gradient descent, we often observe that AC trajectories are step functions with long periods of time between finding improving configurations, complicating the prediction of whether improvements will still happen.
For these reasons and to enable an efficient use of resources, we chose to treat the budgets as discussed in the text.};
when terminated, it returns its current incumbent, i.e., the best found configuration so far.
%

%%%%%%%%%%%%%%%%%%%%%%%%%%%%%%%%%%%%%%%%%%%%%%%%%%%%%%%%%%%%%%%%%%%%
\subsection{Why and When to Consider AC?}
\label{sec:acoutsiderwhy}
%%%%%%%%%%%%%%%%%%%%%%%%%%%%%%%%%%%%%%%%%%%%%%%%%%%%%%%%%%%%%%%%%%%%

Algorithm configuration should always be considered if
(i) the empirical performance of an algorithm is relevant
and (ii) the algorithm has performance-relevant parameters.
This is quite obvious for most empirical studies showing
that a new algorithm $\algo$ establishes a new state-of-the-art performance
on benchmark problem X.
However, in this setting it is also important to tune the parameters
of all algorithms to compare against --- without 
this, a comparison would not be fair because
one of the algorithms may only perform best because its parameters
were tuned with the most effort~\cite{hooker-jh95a}.
Indeed, as shown several times in the AC literature, optimized configurations often perform much better than default ones; in some cases, the default configuration may even be worse
than one drawn uniformly at random 
(e.g., see Figure~\ref{fig:overtuning_instances:rooks}).  

There are several other advantages of AC compared to manual parameter tuning \cite<cf. >{lopez-ibanez-orp16}, including:

\begin{description}
	\item[Reproducibility] Automated algorithm configuration is often more reproducible 
	than doing manual parameter tuning. 
	Manual parameter tuning strongly depends on the experience
	and intuition of an expert for the algorithm at hand and/or for the given instance set.
	This manual procedure can often not be reproduced by other users.
	If algorithm developers also make their configuration spaces available 
	(e.g., as the authors of \lingeling~\cite{biere-sat14a} and \clasp~\cite{gebser-ai12} do),
	reproducing the performance of an algorithm using AC is feasible.
	\item[Less human-time] Assuming that a reasonable configuration space is known,
	applying algorithm configuration is often much more efficient than manual parameter tuning.
	While ceding this tedious task to algorithmic approaches can come at the cost of requiring more computational resources, these tend to be quite cheap compared to paying a human expert and are increasingly widely available.
	\item[More thoroughly tested] Since humans are impatient by nature (e.g., during the development of algorithms),
	they often focus on a rather small subset of instances to get feedback fast 
	and to evaluate another configuration. Compared to humans, configurators often evaluate (promising) configurations
	more thoroughly on more instances.
	\item[More configurations evaluated] Because of similar reasons as above, humans tend to 
	evaluate far less configurations than most configurators would do.
\end{description}

\noindent However, there are also two major limitations of AC, which must be considered:
\begin{description}
	\item[Homogeneous instances] To successfully apply AC,
	the instances have to be similar enough such that 
	configurations that perform well on 
	subsets of them also tend to perform well on others;
	we call such instance sets \emph{homogeneous}.
	If the instances are not homogeneous, it is harder to find a configuration that 
	performs well on average; it is even possible that a configurator returns a configuration $\conf$
	that performs worse than the default one (although $\conf$ may appear to perform better based on the instances the configurator could consider within its limited budget). 
	Unfortunately, so far, none of the existing AC tools implement an automatic check whether
	the given instances are sufficiently homogeneous.
	For heterogeneous instance sets, portfolio approaches~\cite{xu-jair08a,kadioglu-cp11a,malitsky-cp12a,lindauer-jair15a} or instance-specific algorithm configuration~\cite{xu-aaai10a,kadioglu-ecai10} provide alternative solutions.
	\item[Specialization] From the restriction to homogeneous instances,
	the second limitation of AC follows: 
	the optimized configurations (returned by a configurator) are always 
	specialized to the instance set and cost metric at hand.
	%at hand.
	It is hard to obtain a robust configuration on a large variety of heterogeneous instances. (In fact, it is not even guaranteed that a single configuration with strong performance on all instances exists.)
\end{description}

%%%%%%%%%%%%%%%%%%%%%%%%%%%%%%%%%%%%%%%%%%%%%%%%%%%%%%%%%%%%%%%%%%%%
\subsection{Setting up AC Experiments}
\label{sec:acoutsidershow}
%%%%%%%%%%%%%%%%%%%%%%%%%%%%%%%%%%%%%%%%%%%%%%%%%%%%%%%%%%%%%%%%%%%%

In the following, we describe the typical steps to set up and run AC experiments, and provide pointers to the pitfalls and best practices discussed later. 

\begin{enumerate}
  \denselist
	\item Define an instance set of interest, which should be homogeneous (see Section~\ref{sec:homo}) and representative of future instances (see Section~\ref{sec:rep_insts_cutoff});
	\item Split your instances into a training and test instances (see Section~\ref{sec:train_test}); the test instances are later used to safeguard against over-tuning effects (see Section~\ref{sub:over-instances});
	\item Define the ranges of all performance-relevant parameters giving rise to the configuration space (see Sections \ref{sec:config_space} and \ref{sec:which_params});
	\item Implement the interface between your algorithm and the configurator; take Pitfalls 1-4 into consideration (Section~\ref{sec:pitfalls_wrapper});
	\item Choose your preferred configurator (e.g., \paramils, \gga++, \irace{} or \smac; see Section \ref{sec:ac:approaches}) 
	\item Define the resource limitations your algorithm (cutoff time and memory limit) and the configurator (configuration budget) should respect (see Section~\ref{sec:settings});
	\item Define your cost metric to be optimized; if the cost metric is runtime, configurators typically optimize PAR10 as the metric of interest, which is the penalized average runtime (in CPU seconds)
	counting runs exceeding the cutoff time $\cutoff$  as $10 \cdot \cutoff$; furthermore please consider Pitfalls 2 and 3 (see Section~\ref{sub:not_term}) and recommendations in Section~\ref{sec:time_metric} for runtime optimization; if the cost metric is related to the quality of the solution, e.g. the error of a model on a dataset, configurators typically minimize validation error.
	\item Run the AC experiments on the training instances and obtain the final incumbent---consider to use parallel runs (Section~\ref{sec:parallel_res});
	\item Evaluate the default configuration and the optimized configuration on the test instances, to obtain an unbiased estimate of generalization performance on new instances, and to assess over-tuning effects (Section~\ref{sec:pitfalls_overtuning});
	\item Optionally, use further tools to obtain visualizations and gain more insights from the AC experiments, e.g., CAVE~\cite{biedenkapp-lion18}. 
\end{enumerate}

\begin{figure}
\centering
\includegraphics[width=0.4\textwidth]{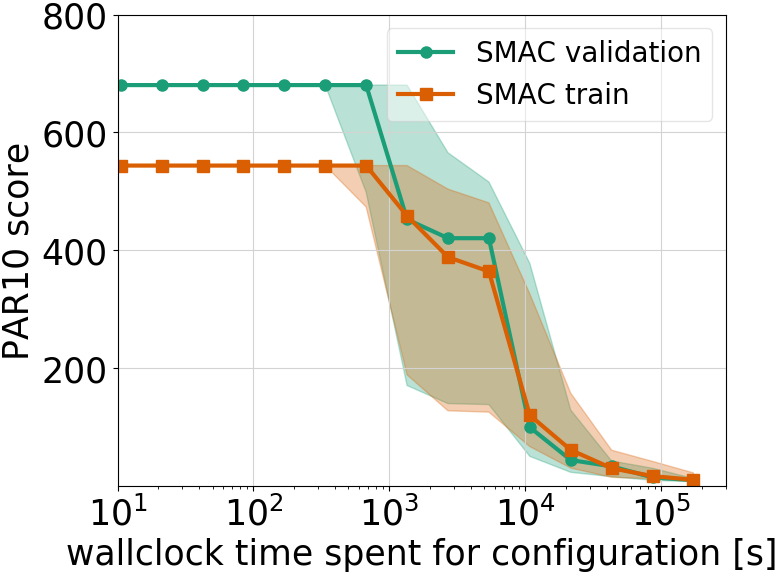}
\caption{Exemplary application of AC, optimizing 75 parameters of \clasp{} to solve N-Rooks problems. At each time step $t$, we show the penalized average runtime (PAR10) score on the training set (green) and test set (orange) of the incumbent configuration at time $t$. I.e., at each time step, we take the best configuration found so far (the one the configurator would return if stopped at that time), ran the algorithm with it on the training and test set and recorded its PAR10 score. We show the median and quartiles of repeating this process 16 times using different random seeds.\label{fig:ACShowcase}}
\end{figure}

As an exemplary application where AC yields dramatic speedups, we ran \smac{} to optimize $75$ parameters of the configurator \clasp~\cite{gebser-ai12} to solve N-Rooks~\cite{manthey-sat14r} instances. We will return to this scenario in more detail in Subsection~\ref{sub:over-instances}. Here, we used a training set of $484$ instances and a test set of $351$ instances to evaluate the best found configurations over time. 
We used a cutoff of $300$ seconds, within which the default configuration solves $82\%$ of all training instances. 
Figure~\ref{fig:ACShowcase} reports results from $16$ independent \smac{} runs, showing that AC using an adequate setup can robustly yield large speedups compared to not tuning the algorithm.

%%%%%%%%%%%%%%%%%%%%%%%%%%%%%%%%%%%%%%%%%%%%%%%%%%%%%%%%%%%%%%%%%%%%
\subsection{Approaches for Solving the AC problem} \label{sec:ac:approaches}
%%%%%%%%%%%%%%%%%%%%%%%%%%%%%%%%%%%%%%%%%%%%%%%%%%%%%%%%%%%%%%%%%%%%

For subproblems of the AC problem that deal neither with instances
nor with capped and censored runs, there exist several approaches in the fields of
parameter tuning, hyperparameter optimization and expensive black-box optimization.
Prominent examples include Bayesian optimization~\cite{mockus-tgo78a,shahriari-ieee16a},
sequential parameter optimization~\cite{bartzbeielstein-emaoa10a}, evolution strategies~\cite{hansen-eda06}, 
and combinations of several classical search strategies~\shortcite{ansel-pact14a}.

For solving the full AC problem, there are several configurators. 
\paramils{}~\cite{hutter-jair09a} uses local search in the configuration space, employing a racing strategy to decide which of two configurations performs better without running both of them on all instances.
Recently, \citeA{caceres-endm17a} also proposed to use variable neighborhood search instead of the iterated local search used in \paramils.
\irace{}~\cite{lopez-ibanez-orp16} uses iterative races via F-race~\cite{birattari-gecco02a} on a set of sampled
%picked
 configurations to determine the best one.
\smac{}~\cite{hutter-lion11a} and its distributed version
\dsmac{}~\cite{hutter-lion12a} use probabilistic models of algorithm
performance, so-called empirical performance models~\cite{hutter-aij14a}, to guide the search for good configurations by means of an extension of Bayesian Optimization~\cite{brochu-arXiv10a}.
\gga{}~\cite{ansotegui-cp09a} represents parameters as genes and uses a genetic algorithm with a competitive and a non-competitive gender; its newest version \gga{}++~\cite{ansotegui-ijcai15a} also uses an empirical performance model for guidance. For a more detailed description of these algorithms, we refer the interested reader to the original papers or to the report of the Configurable SAT Solver Challenge~\cite{hutter-aij17a}. 

If the cost metric $c$ is \runtime{} using PAR10 scores,
several configurators use an adaptive capping
strategy~\cite{hutter-jair09a} to terminate slow algorithm runs prematurely
to save time.\footnote{As a side note, we remark that for model-based methods the internal model needs to handle dynamic timeouts arising from adaptive capping and PAR10 scores for guiding the search are based on predictions of that model. Furthermore, evaluations of incumbents for validation purposes are done purely with a fixed timeout $\cutoff_{max}$, making PAR10 values comparable across configurators.}
For example, if the maximal cutoff time used at test time is $\cutoff_{max} = 5000$
seconds and the best configuration known so far solves each instance in 10
seconds, we can save dramatically by cutting off slow algorithm runs
after $\cutoff>10$ seconds instead of running all the way to $\cutoff_{max}$.
Since $\cutoff$ is adapted dynamically, each target algorithm
run can be issued with a different one.

%%%%%%%%%%%%%%%%%%%%%%%%%%%%%%%%%%%%%%%%%%%%%%%%%%%%%%%%%%%%%%%%%%%%
\subsection{The Role of the Target Algorithm Wrapper}
\label{sec:wrapper}
%%%%%%%%%%%%%%%%%%%%%%%%%%%%%%%%%%%%%%%%%%%%%%%%%%%%%%%%%%%%%%%%%%%%

As depicted in Figure \ref{fig:ac}, configurators execute the target algorithm with configurations $\conf \in \pcs$ on instances $\pi \in \Pi$ and measure the resulting cost $c(\conf, \pi)$. To be generally applicable, configurators specify an interface through which they evaluate the cost $c(\conf, \pi)$ of arbitrary algorithms to be optimized.
For a new algorithm $\algo$, users need to implement this interface to actually
execute $\algo$ with the desired configuration $\conf$ on the desired instance
$\inst$ and measure the desired cost metric $c(\conf, \pi)$ 
(e.g.~runtime required to solve a SAT instance or validation error of a machine learning model).

In order to avoid having to change the algorithm to be
optimized, this interface is usually implemented by a
\emph{wrapper}.\footnote{An alternative to a general wrapper would be
a programming language-specific reliable interface for the communication between
configurator and target algorithm~\cite{hoos-cacm12}, which would make it
easier for users to apply algorithm configuration to new target algorithms.
However, the design of such an interface would also need to consider the
pitfalls identified in this paper.} In the simplest case, the input to the
wrapper is just a parameter configuration $\conf$, but in general AC it also includes an instance $\inst$, and it can also include a random seed and computational resource limits, such as a \runtime{} cutoff $\cutoff$.
Given these inputs, the wrapper executes the target algorithm with configuration $\conf$ on instance $\inst$, and measures and returns the desired cost metric $c(\conf, \pi)$.

%%%%%%%%%%%%%%%%%%%%%%%%%%%%%%%%%%%%%%%%%%%%%%%%%%%%%%%%%%%%%%%%%%%%
\section{Pitfalls and Best Practices Concerning Algorithm Execution}
\label{sec:pitfalls_wrapper}
%%%%%%%%%%%%%%%%%%%%%%%%%%%%%%%%%%%%%%%%%%%%%%%%%%%%%%%%%%%%%%%%%%%%

In this and the next section, we describe common pitfalls in algorithm configuration
and illustrate their consequences on existing benchmarks from the algorithm configuration library \aclib~\shortcite{hutter-lion14a}\footnote{See \url{www.aclib.net}}.
Based on the insights we acquired in thousands of algorithm configuration experiments over the years, we propose best practices to avoid these pitfalls. 

Throughout, we will use the state-of-the-art configurator
\smac{}~\cite{hutter-lion11a} as an example, typically optimizing PAR10.
Where not specified otherwise, we ran all experiments on the University of Freiburg's META cluster, each of whose nodes shares 64 GB of RAM among two Intel Xeon E5-2650v2 8-core CPUs with 20 MB L3 cache and runs Ubuntu 14.04 LTS 64 bit.\footnote{Data and scripts for the experiments in this paper are available at \\ \url{http://www.automl.org/best-practices-in-algorithm-configuration/}.}
 
%%%%%%%%%%%%%%%%%%%%%%%%%%%%%%%%%%%%%%%%%%%%%%%%%%%%%%%%%%%%%%%%%%%%
\subsection{\emph{Pitfall 1: Trusting Your Target Algorithm}}
\label{sub:trust_sol}
%%%%%%%%%%%%%%%%%%%%%%%%%%%%%%%%%%%%%%%%%%%%%%%%%%%%%%%%%%%%%%%%%%%%

Many state-of-the-art algorithms have been exhaustively benchmarked and tested
with their default parameter configuration.
However, since the configuration space of many algorithms is very large, we
frequently observed hidden bugs triggered only by rarely-used combinations of
parameter values. For example, \citeA{hutter-cpaior10a} reported finding bugs
in mixed integer programming solvers and \citeA{manthey-sat16a} bugs in SAT solvers.
Due to the size of the associated configuration spaces \cite<e.g., 214 parameters 
and a discretized space of $10^{86}$ configurations 
in the state-of-the-art SAT solver \textit{Riss},>{riss}, exhaustive checks are infeasible in practice.

Over the years, the types of bugs we have experienced even in commercial solvers
(that are the result of dozens of person-years of development time) include:
\begin{itemize}
\denselist
	\item Segmentation faults, Null pointer exceptions, and other unsuccessful algorithm terminations;
	\item Wrong results (e.g., claiming a satisfiable SAT instance to be unsatisfiable);
	\item Not respecting a specified \runtime{} cutoff that is passed as an input;
	\item Not respecting a specified memory limit that is passed as an input;
	\item Rounding down \runtime{} cutoffs to the next integer (even if that integer is zero); and
	\item Returning faulty \runtime{} measurements (even negative ones!)
\end{itemize}

\paragraph{Effects}
The various issues above have a multitude of negative effects, from obvious to subtle. 
If the algorithm run does not respect its resource limits this can lead to
congested compute nodes (see Pitfall 3) and to configurator runs that are stuck
waiting for an endless algorithm run to finish. Wrongly reported \runtime{}s
(e.g., close to negative infinity in one example) can lead to endless
configuration runs when trusted. Rounding down cutoff times can let
configurators miss the best configuration (e.g., when they use adaptive capping
to cap \runtime{}s at the best observed \runtime{} for an instance -- if
that \runtime{} is below one second then each new configuration will fail on
the instance due to using a cutoff of zero seconds).

Algorithm crashes can be fairly benign when they are noticed and counted with
the highest possible cost, but they can be catastrophic when not recognized as
crashes: e.g., when blindly minimizing an algorithm's \runtime{} the
configurator will typically simply find a configuration that crashes quickly.
While this can be exploited to quickly find
bugs~\cite{hutter-cpaior10a,manthey-sat16a}, obtaining faulty
configurations is typically the worst possible result of using
algorithm configuration in practice.
Bugs that lead to wrong results tend to be discovered by configurators when
optimizing for \runtime{}, since (at least for $\mathcal{NP}$-hard
problems) we found that such bugs often allow algorithms to find shortcuts
and thus shorten \runtime{}s. Therefore, blindly minimizing \runtime{}
without solution checking often yields faulty configurations.

\paragraph{Detailed Example}
In 2012, we used algorithm configuration to minimize the \runtime{} of the state-of-the-art solver \textit{glucose}~\cite{audemard-ijcai09a}.
We quickly found a parameter configuration that appeared to yield new state-of-the-art performance on the industrial instances of the SAT Challenge 2012\footnote{\url{http://baldur.iti.kit.edu/SAT-Challenge-2012/}}; however, checking this configuration with the authors of \glucose{} revealed that it led to a bug which made \glucose{} falsely report some satisfiable instances as unsatisfiable.\footnote{The bug in \glucose{} version 2.1 was fixed after we reported it to the developers, and we are not aware of any bugs in the newest \glucose{} version~4.1.}

\begin{figure}[t]
\centering
\includegraphics[width=0.7\textwidth]{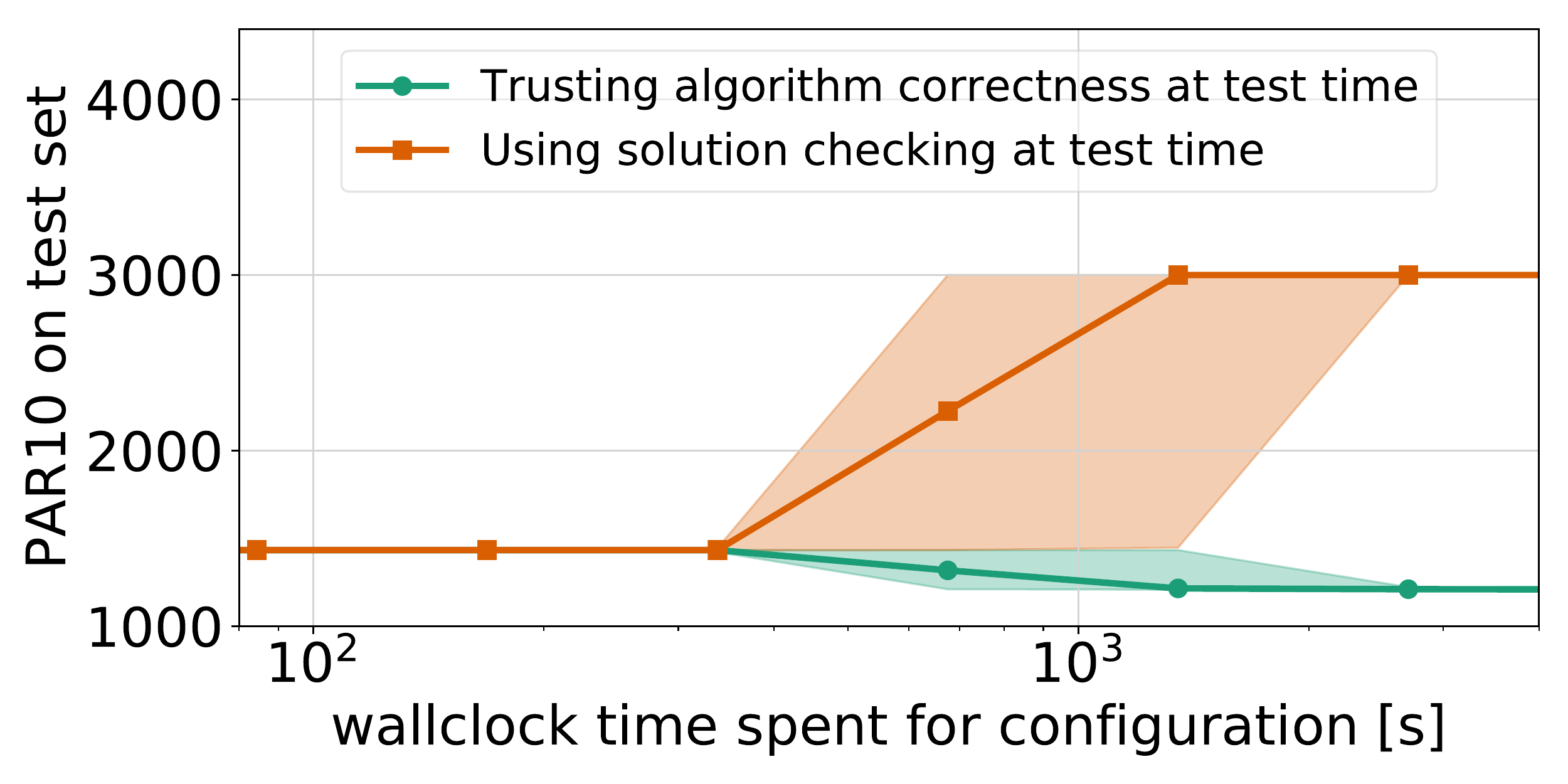}
\caption{Difference in test set performance as judged when trusting the target
algorithm (green) and using external solution checking (orange). 
We plot the penalized average \runtime{} (PAR10) scores of \glucose{} v2.1 on
the industrial instances from the SAT Challenge 2012, as a function of time
spent for configuration, when the configuration process trusted \glucose{} v2.1
to be correct.
We ran $12$ \smac{} runs and at each time step show the median and quartiles of
their incumbents' scores. The green curve computes these scores trusting the
solutions \glucose{} returns, while the orange curve penalizes faulty
configurations with the worst value of 3000 (where faulty configurations are
those that yield at least one wrong result on the test instances; such
configurations would, e.g., be disqualified in the SAT competition).
We emphasize that both curves are based on exactly the same set of 12 SMAC runs (which were broken in that they trusted \glucose{} rather than applying solution checking) and only differ in their validation.
}
\label{fig:glucose:bug}
\end{figure}

In Figure~\ref{fig:glucose:bug} we reconstruct this behaviour. We ran \smac{} on \glucose{} v2.1 and evaluated configurations found over time when trusting \glucose{}'s correctness at configuration time: The green curve shows \glucose{}’s (buggy) outputs on the test instances, whereas the orange curve scored each configuration using solution checking, and returning the worst possible score for configurations that returned a wrong solution. 
After 300 to 3000 seconds, \smac{} found configurations that seemed better when trusting \glucose{}’s outputs, but that actually sometimes returned wrong solutions, resulting in the true score (orange curve) going up (getting worse) to the worst possible PAR10 score.

\paragraph{Best Practice}

Most of the issues above can be avoided by wrapping target algorithm runs with a reliable piece of code that limits their resources and checks whether they yield correct results. Cast differently, the job of this wrapper is to actually measure the cost function $c(\conf,\inst)$ of interest, which should intuitively heavily penalize any sort of crashes or bugs that lead to wrong results. 

If enough computational time is available, we recommend to first run systems such as \textit{SpyBug}~\cite{manthey-sat16a} to find bugs in the configuration space, and to either fix them or to exclude the faulty part of the configuration space from consideration.
Regardless of whether this is done or not, since it is infeasible to perfectly
check the entire configuration space, we always recommend to check the returned
solution of the target algorithms during the configuration process. For example,
for SAT instances, our example wrapper exploits the
standard SAT checker tool routinely used in the SAT competitions to verify the
correctness of runs. For solvers that output unsatisfiability proofs, there are also effective tools for checking these proofs~\cite{heule-stvr14a}.

%%%%%%%%%%%%%%%%%%%%%%%%%%%%%%%%%%%%%%%%%%%%%%%%%%%%%%%%%%%%%%%%%%%%
\subsection{\emph{Pitfall 2: Not Terminating Target Algorithm Runs Properly}}\label{sub:not_term}
%%%%%%%%%%%%%%%%%%%%%%%%%%%%%%%%%%%%%%%%%%%%%%%%%%%%%%%%%%%%%%%%%%%%

Given the undecidability of the halting problem, target algorithm runs need to
be limited by some kind of \runtime{} cutoff $\cutoff_{max}$ to prevent poor
configurations from running forever.
In many AI communities, it is a common practice to set a \runtime{} cutoff as
part of the cost metric and measure the number of timeouts with that cutoff
(e.g., $\cutoff_{max}=5000$ seconds in the SAT race series).
In algorithm configuration, the ability to prematurely cut off unsuccessful
runs also enables adaptive capping (see Section~\ref{sec:ac}). Therefore, it is
essential that target algorithm runs respect their cutoff.
This pitfall is related to Pitfall 1 as the user also needs to trust the target algorithm to work appropriately. While for Pitfall 1 we focus on the returned solution, here we draw attention to the resource limitations.

\paragraph{Effect}

Consequences of target algorithm runs not respecting their cutoffs can include:

\begin{enumerate}
  \item If the target algorithm always uses the maximal cutoff $\cutoff_{max}$
  and ignores an adapted cutoff $\cutoff < \cutoff_{max}$, the configuration
  process is slowed down since the benefits of adaptive capping are given up;  
  \item If the target algorithm completely ignores the cutoff, the configuration
  process may stall since the configurator waits for a slow target algorithm
  to terminate (which, in the worst case, may never happen);
  \item If a wrapper is used that fails to terminate the actual algorithm run
  but nevertheless returns the control flow to the configurator after the
  cutoff time $\cutoff$, then the slow runs executed by the configurator will
  continue to run in parallel and overload the machine, messing up the cost
  computation (e.g., wallclock time).
\end{enumerate}

\paragraph{Example}

The latter (quite subtle) issue actually happened in a recent publication that compared \gga{}++ and \smac{}, in which a wrapper bug caused \smac{} to perform poorly \cite{ansotegui-ijcai15a}.
The authors wrote a wrapper for \smac{} that tried to terminate its target algorithm runs (here:
\glucose{} or \lingeling) after the specified cutoff time $\cutoff$ by sending a
KILL signal, but since it ran the target algorithm through a shell (using
\texttt{subprocess.Popen(cmd, shell=True)} in Python) the KILL signal only
terminated the shell process but not the actual target algorithm (which
continued uninterrupted until successful, sometimes for days).
When attempting to reproduce the paper's experiments with the original wrapper kindly
provided by the authors, over time more and more target algorithms were spawned
without being terminated, causing our 16-core machine to slow down and
eventually become unreachable.
This issue demonstrates that \smac{} heavily relies on a robust wrapper that automatically terminates its target algorithm runs properly.\footnote{In contrast to \smac{}, \gga{}++ does not require a wrapper; in the experiments by \citeA{ansotegui-ijcai15a}, \gga{}++ directly sent its KILL signal to the target algorithm and therefore did not suffer from the same problem \smac{} suffered from, which confounded the paper's comparison between \gga{}++ and \smac{}. 
Additionally, there was also a simple typo in the authors' wrapper for \smac{} in parsing the target
algorithm's output (here: \glucose{}) that caused it to count all successful
runs on unsatisfiable instances as timeouts. Receiving wrong results for all
unsatisfiable instances (about half the instance set) severely affected \smac{}'s trajectory; this issue was only present in the wrapper for \smac{} (and therefore did
not affect \gga{}++), confounding the  
comparison between \gga{}++ and \smac{} further.}

\begin{figure}[t]
\centering
\includegraphics[width=0.7\textwidth]{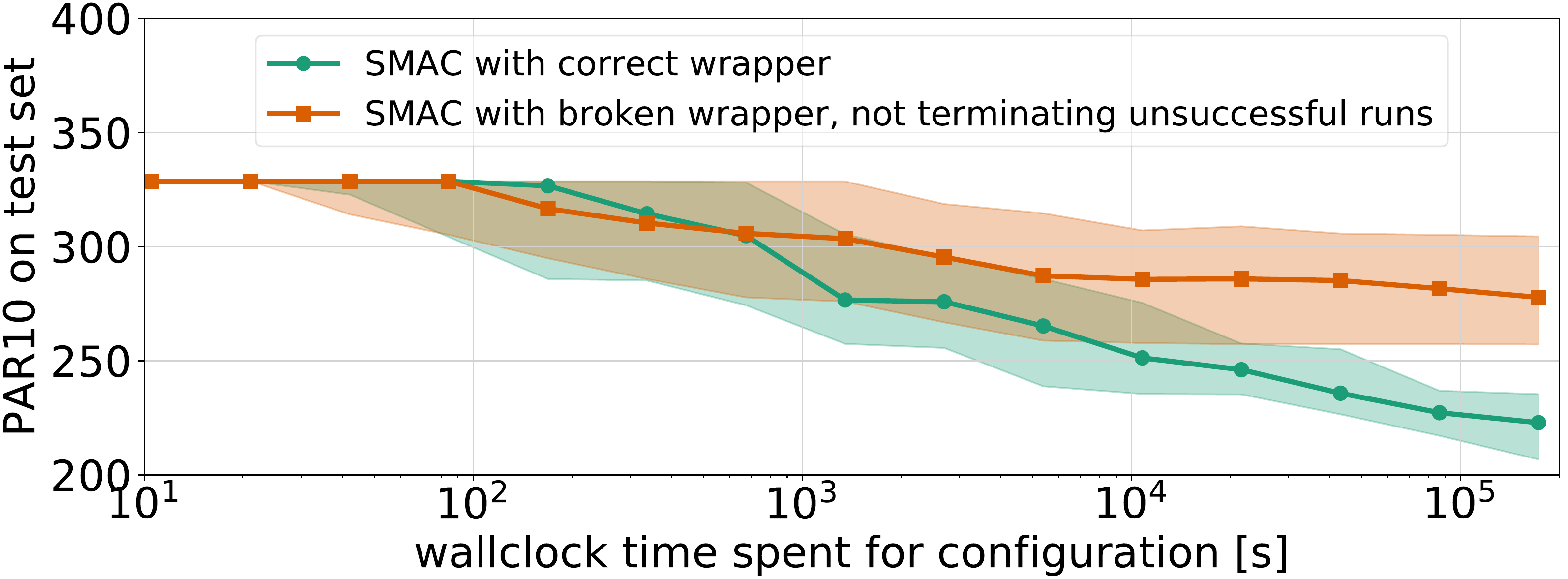}
\caption{Effect of a broken wrapper that does not terminate target algorithm
runs properly. We show PAR10 test set performance for optimizing
\cryptominisat{} with \smac{} on \circuitfuzz{} instances, when using a correct and a
broken wrapper during configuration, respectively. We show median test
performance (measured using a correct wrapper) with quartiles across 80 runs of
\smac{}. Not terminating target algorithm runs properly eventually slowed
down the machine affecting \runtime{} measurements.
\label{fig:wrapperbug}}
\end{figure}
To illustrate this issue in isolation, we compared \smac{} using a working wrapper and a broken version of it that returns the control flow to the configurator when the \runtime{} cutoff is reached, without terminating the target algorithm run process.
Figure~\ref{fig:wrapperbug} shows the performance achieved when \smac{} is run with either wrapper to configure \cryptominisat{}~\cite{cryptominisat} for penalized average \runtime{} (PAR10) to solve \circuitfuzz{} instances~\cite{brummayer-sat10a}
as used in the CSSC 2014~\cite{hutter-aij17a}.
We executed $80$ \smac{} runs for each wrapper, with $16$ independent parallel runs each on five 16-core machines. Both \smac{} versions performed equally well
until too many target algorithm processes remained on the machines and
prevented \smac{} from progressing further.
Only on one of the five machines that ran \smac{} with the broken
wrapper, the runs terminated after the specified wallclock-limit of 2 days;
after an additional day, three of the remaining machines were still frozen caused by overload and the fourth could not be reached at all.

\paragraph{Best Practice}

To avoid this pitfall, we recommend to use some well-tested, external piece of code to reliably control and terminate target
algorithm runs.

%%%%%%%%%%%%%%%%%%%%%%%%%%%%%%%%%%%%%%%%%%%%%%%%%%%%%%%%%%%%%%%%%%%%
\subsection{\emph{Pitfall 3: Slow File System}}\label{sub:slow_file}
%%%%%%%%%%%%%%%%%%%%%%%%%%%%%%%%%%%%%%%%%%%%%%%%%%%%%%%%%%%%%%%%%%%%

Related to Pitfall 2, another way to ruin \runtime{} measurements by slowing
down a machine is to overload the used file system. 
Each target algorithm run typically has to read the given problem instance and
writes some log files; thus, executing many algorithm configuration runs in
parallel can stress the file system.

\paragraph{Effect}
Slowdowns caused by an overloaded file system can have a severe impact on runtime measurements; in particular this is problematic because most algorithm configurators measure their own configuration budget as wallclock time. Furthermore, these problems are often not immediately recognizable (because everything runs fine when tested at small scale) and sometimes only affect parts of a large set of experiments (as the overload might only happen for a short time).

\paragraph{Example 1}
Over the years, we have experienced file system issues on a variety of clusters
with shared file systems when target algorithm runs were allowed to write
to the shared network file system. When executing hundreds (or on one cluster,
even thousands) of algorithm configuration runs in parallel, this stressed the
file system to the point where the system became very slow for all users and we
measured 100-fold overheads in individual target algorithm evaluations. Writing
target algorithm outputs to the local file system fixed these issues.

\paragraph{Example 2}
Distributing configuration runs across multiple nodes in a compute
cluster (e.g., in \gga{}, \irace{}, or \dsmac{}) can be error-prone if the configurators
communicate via the file system.
In particular, we experienced issues with several shared network file systems
with asynchronous I/O; e.g., on one compute node a file was written, but that
file was not immediately accessible (or still empty) on other compute
nodes.
Often a second read access resolved the problem, but this solution can be
brittle; a change of parallelization strategy may in that case yield more robust
results.

\paragraph{Example 3}
Even when writing target algorithm output to the local file system, we once experienced
200-fold overheads in target algorithm runs (invocations of sub-second target
algorithm runs hanging for minutes) due to a subtle combination of issues when
performing hundreds of algorithm configuration experiments in parallel.
On the Orcinus cluster (part of Compute Canada's Westgrid cluster), which
uses a Lustre file system, we had made our algorithm configuration benchmarks 
read-only to prevent accidental corruption.
While that first seemed like a good idea, it disallowed our Python wrapper to
create \texttt{.pyc} bytecode files and forced it to recompile at every
invocation, which in turn triggered a stats call (similar to
\texttt{ls} on the Linux command line) for each run.
Stats calls are known to be slow on the Lustre file system, and executing
them for each sub-second target algorithm run on hundreds of compute nodes in parallel led
to extreme file system slowdowns. After testing many other possible reasons for
the slowdowns, removing the read-only condition immediately fixed all issues.

\paragraph{Best Practice}
Issues with shared file systems on compute clusters can have subtle
reasons and sometimes require close investigation (as in our Example 3). 
Nevertheless, most issues can be avoided by using the faster local file system
(typically \texttt{/tmp/}, or even better, a
temporary job-specific subdirectory thereof\footnote{We note that on some modern
Linux distributions, \texttt{/tmp/} can be a RAM disk and therefore may use
resources allotted to the algorithm runs; in general, we recommend to make
the choice about a fast temporary directory specific to the compute cluster used.}) for
all temporary files, and by measuring CPU time instead of wallclock time (at least for sequential algorithms).

%%%%%%%%%%%%%%%%%%%%%%%%%%%%%%%%%%%%%%%%%%%%%%%%%%%%%%%%%%%%%%%%%%%%
%\subsection*{Pitfall 4: Comparing Configurators using Different Wrappers}
\subsection{\emph{Pitfall 4: Handling Target Algorithm Runs Differently}}
\label{sub:different_wrappers}
%%%%%%%%%%%%%%%%%%%%%%%%%%%%%%%%%%%%%%%%%%%%%%%%%%%%%%%%%%%%%%%%%%%%

The required functionalities of the target algorithm wrapper differ slightly for
different configurators.
For example, \smac{} and \paramils{} trust the wrapper to terminate target
algorithms, but \gga{} sends a KILL signal on its own~(see also Pitfall 2).
Therefore, sometimes configurators are compared by using different target algorithm calls and measurements. 
However, if this is not done properly, it can lead to a biased comparison between configurators.

\paragraph{Effect}

Calling the target algorithm differently for different configurators can lead to different
behaviors of the target algorithm and hence, to different returned performance
values for the same input.
If the configurators receive different performance measurements, they will
optimize different objective functions and their runs become incomparable.

\paragraph{Example}

During the early development of \smac{} (before any publication), we used the same wrappers for
\paramils{} and \smac{} but an absolute path to the problem instance for one and
a relative path for the other. Even this tiny difference lead to reproducible
differences of \runtime{} measurements of up to 20\% when optimizing an algorithm
implemented in UBCSAT 1.1.0~\cite{tompkins-sat05a}.
The reason was that that version of UBCSAT stored its callstring in its heap
space such that the number of characters in the instance name affected
data locality and therefore the number of cache misses and the \runtime{}
(whereas the number of search steps stayed the same).\footnote{This issue is fixed in later versions of UBCSAT.}
This subtle issue demonstrates the importance of using the same wrapper
for all configurators being compared such that 
exactly the same target algorithm calls are used.

\paragraph{Best Practice}
We recommend to use a single wrapper when comparing configurators against each
other, in order to guarantee that all configurators optimize the same objective.
For studies comparing configurators, it is also paramount to use
tried-and-tested publicly available benchmark scenarios (lowering the
risk of typos, etc; see also Footnote 9); our algorithm configuration benchmark library
AClib~\cite{hutter-lion14a} provides a very broad collection of such benchmarks.

%%%%%%%%%%%%%%%%%%%%%%%%%%%%%%%%%%%%%%%%%%%%%%%%%%%%%%%%%%%%%%%%%%%%
\section{Pitfalls and Best Practices Concerning Over-Tuning}
\label{sec:pitfalls_overtuning}
%%%%%%%%%%%%%%%%%%%%%%%%%%%%%%%%%%%%%%%%%%%%%%%%%%%%%%%%%%%%%%%%%%%%

A common issue in applying algorithm configuration is the over-tuning
effect~\cite{birattari-PhD2004,hutter-aaai07a,birattari-tuning09a,hutter-jair09a}
Over-tuning is very related to the concept of over-fitting in machine learning
and denotes the phenomenon of finding parameter configurations that
yield strong performance for the training task but do not generalize to test
tasks.
We emphasize that over-tuning effects are not necessarily only related to the set of training instances used, but can also include the characteristics of the experimental setup such as the resource limitations and bugs in the solver (see Pitfall 1).
To safeguard against over-tuning effects, it is crucial to evaluate
generalization performance (typically, using a set of benchmark instances
disjoint from the benchmarks used for training).
In the following, we discuss three pitfalls related to over-tuning.

%%%%%%%%%%%%%%%%%%%%%%%%%%%%%%%%%%%%%%%%%%%%%%%%%%%%%%%%%%%%%%%%%%%%
\subsection{\emph{Pitfall 5: Over-tuning to Random Seeds}}
\label{sub:random}
%%%%%%%%%%%%%%%%%%%%%%%%%%%%%%%%%%%%%%%%%%%%%%%%%%%%%%%%%%%%%%%%%%%%

Many algorithms are randomized (e.g., SAT solvers or AI planners).
However, in many communities, the random seeds of these algorithms 
are fixed to simulate a deterministic behavior
and to ensure reproducibility of benchmark results.

\paragraph{Effect}

\begin{figure}
\begin{tabular}{cc}
\multicolumn{2}{c}{\includegraphics[width=0.8\textwidth]{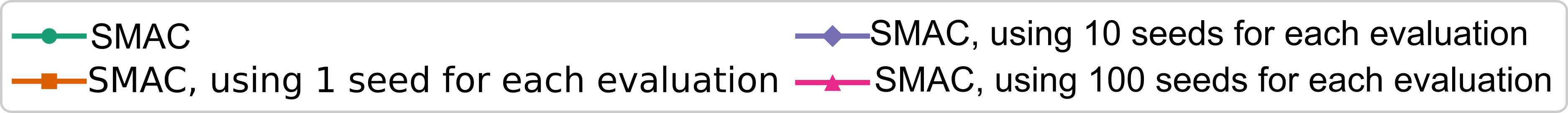}} \\
\includegraphics[width=0.48\textwidth]{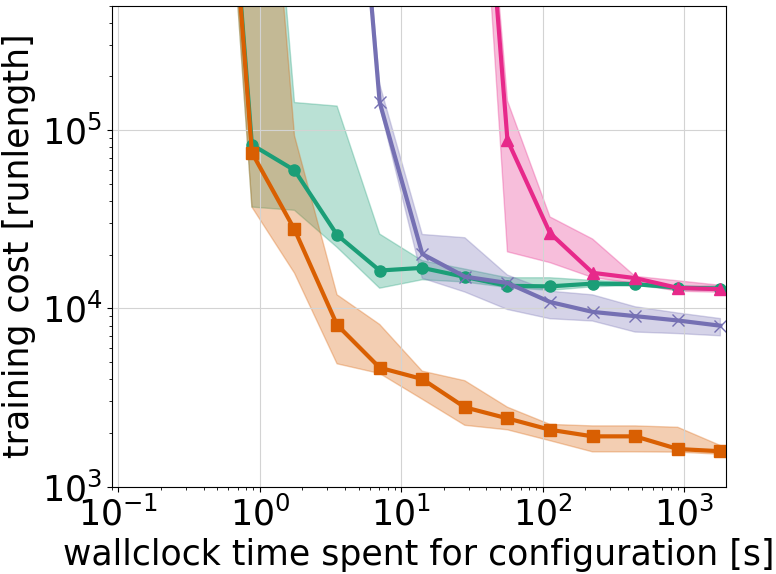} &
\includegraphics[width=0.48\textwidth]{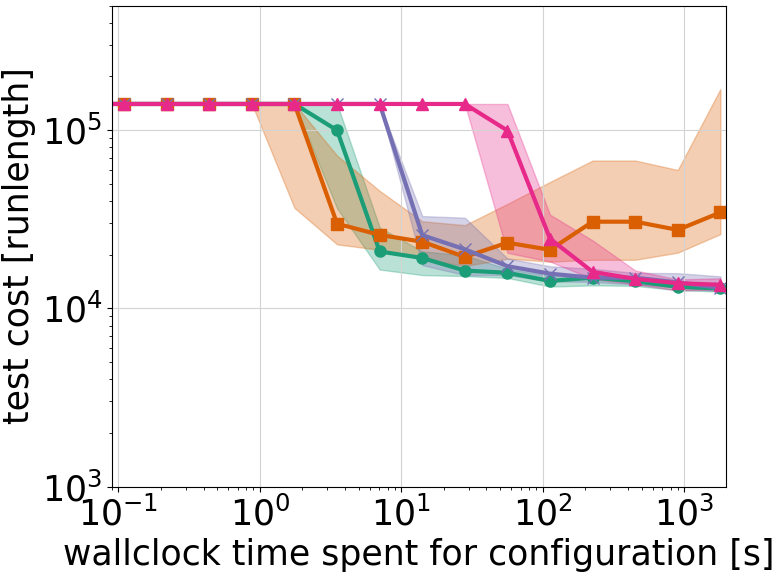}
\end{tabular}
\caption{Optimizing \saps{} with \smac{} on one ``quasigroup with holes'' instance (QWH) and different numbers of random seeds. Each line in the left plot shows the median and quartiles of the estimated training cost and each line in the right plot shows the median and quartiles of the test cost of \saps{} over time across 10 runs of~\smac{}.
}
\label{fig:over-tuning_seed}
\end{figure}

Ignoring the stochasticity of an algorithm in algorithm configuration by fixing
the random seed can lead to over-tuning effects to this seed, i.e., finding a
configuration that yields good performance with this fixed random seed (or set
of seeds) but poor performance when used with other random seeds.
The extreme case is not to only fix the random seed, but to tune the random seed,
which can lead to an even stronger over-tuning effect.\footnote{We note that, in principle, one could construct situations where fixing or even optimizing the seed could lead to good performance if that seed is used in all future experiments and a large number of instances is available to obtain generalization to other instances. However, we believe that the potential misuse of tuning seeds outweighs any potential benefits.}

\paragraph{Example}

To illustrate over-tuning to a random seed in its purest form, independent of a
difference between training and test instances, we optimized the parameters of
the local-search SAT solver \saps~\cite{hutter-cp02a} on a single
instance, the only difference between training and test being the set of
random seeds used.
We used different settings of \smac{} to handle random seeds:
We compared \smac{} using a fixed set of 1, 10 or 100 random seeds for each target algorithm run and standard \smac{}, which handled the random seed itself (using a larger number of seeds to evaluate the best configurations).

As a cost metric, we minimized the average number of local search steps (the
solver's so-called \textit{runlength}) since this is perfectly reproducible.
For the parameter configurations recommended at each step of each \smac{} run,
we measured \smac{}'s training cost (as the mean across the respective sets of seeds
discussed above) as well as its test cost (the mean runlength across 1000
fixed random seeds that were disjoint from the sets of seeds used for
configuration)~\footnote{Note that~\citeA{hutter-aaai07a} used the \emph{median} to aggregate across the 1000 seeds, resulting in slightly lower training and test runlengths.}.

Figure~\ref{fig:over-tuning_seed} shows median costs across
$10$ \smac{} runs, contrasting training cost (left) and test cost (right).
On training, \smac{}, using 1 seed per evaluation quickly improved and achieved the best training cost on its one random seed, but its performance does not generalize to the test seeds. \smac{}, using 10 or 100 seeds per evaluation were slower but generalized better, and standard \smac{} was both fast and generalized best by adaptively handling the number of seeds to run for each configuration.

\paragraph{Best Practice}

For randomized algorithms, we recommend to tune parameter configurations across
different random seeds---most configurators will take care of the required
number of random seeds if the corresponding options are used.
If a configuration's performance does not even generalize well to new random
seeds, we expect it to also not generalize well to new instances.
Furthermore, the number of available instances is often restricted,
but there are infinitely many random seeds which can be easily sampled.
Likewise, when there are only few test instances, at validation time we
recommend to perform multiple runs with different random seeds for each test instance.

%%%%%%%%%%%%%%%%%%%%%%%%%%%%%%%%%%%%%%%%%%%%%%%%%%%%%%%%%%%%%%%%%%%%
\subsection{\emph{Pitfall 6: Over-tuning to Training Instances}}
\label{sub:over-instances}
%%%%%%%%%%%%%%%%%%%%%%%%%%%%%%%%%%%%%%%%%%%%%%%%%%%%%%%%%%%%%%%%%%%%

The most common over-tuning effect is over-tuning to the set of training
instances, i.e., finding configurations that perform well on training
instances but not on new unseen instances.
This can happen if the training instances are not representative for the test
instances; in particular this is often an issue if the training instance set is
too small or the instances are not
homogeneous~\cite{hutter-amai10,hoos-lion12a}, i.e., if there exists no
single configuration with strong performance for all instances. 

\paragraph{Effect}

In practice, over-tuned configurations that only perform well on a small finite set of instances are of little value, because users are typically interested in configurations that also perform well on new instances.
Phrasing this more generally, research insights should also generalize to experiments with similar characteristics.

\paragraph{Example}

\begin{figure}
\begin{tabular}{ccc}
\multicolumn{3}{c}{\includegraphics[width=0.7\textwidth]{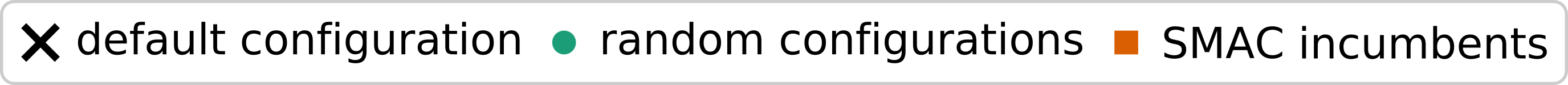}} \\

	\begin{subfigure}[c]{0.3\textwidth}
	\includegraphics[width=1\textwidth]{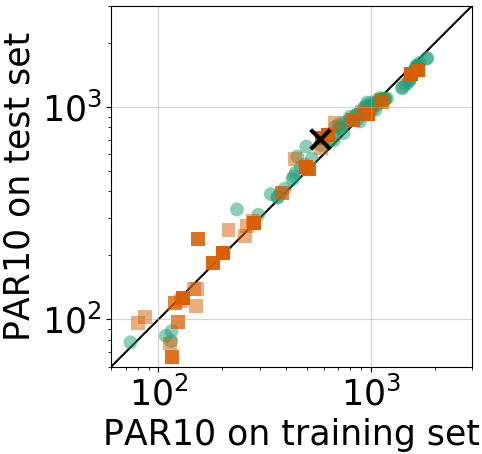}
	\subcaption{\clasp{} on N-Rooks}
	\label{fig:overtuning_instances:rooks}
	\end{subfigure}
	&
	\begin{subfigure}[c]{0.3\textwidth}
	\includegraphics[width=1\textwidth]{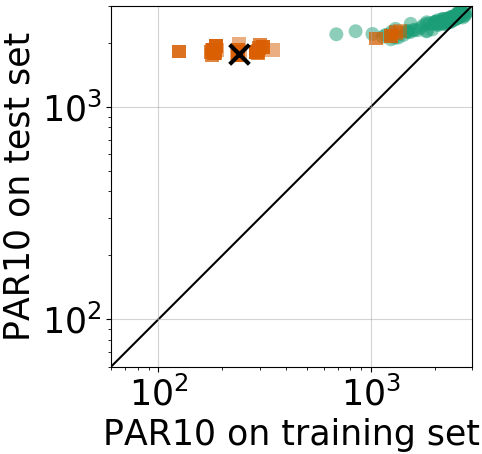}
	\subcaption{\lingeling{} on mixed SAT}
	\label{fig:overtuning_instances:lingeling}
	\end{subfigure}
	&
	\begin{subfigure}[c]{0.3\textwidth}
	\includegraphics[width=1\textwidth]{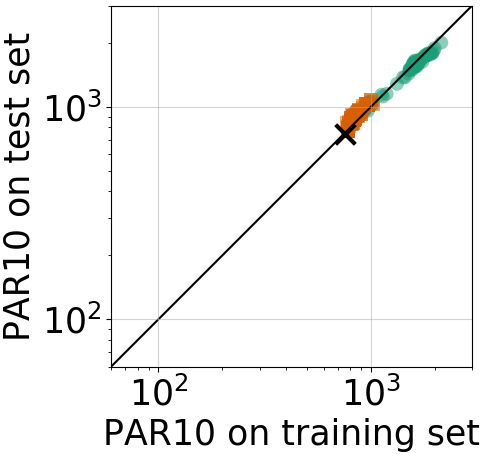}
	\subcaption{\clasp{} on LABS}
	\label{fig:overtuning_instances:labs}
	\end{subfigure}
\end{tabular}
	\caption{Comparing training and test performance of different configurations to study whether these performances on both sets are correlated. Green dots indicate randomly sampled configurations, the black cross marks the performance of the default configuration of the solver, and orange squares correspond to incumbent configurations of 16 \smac{} runs.}
	\label{fig:overtuning_instances}
\end{figure}

To illustrate this problem, we studied training and test performance of
various configurations for three exemplary benchmarks (see Figure~\ref{fig:overtuning_instances}):

\begin{description}
	\item[\clasp{} on N-Rooks] We studied the \runtime{} of the solver
	\clasp{}~\cite{gebser-ai12} on N-Rooks
	instances~\cite{manthey-sat14r}, a benchmark from the Configurable SAT
	Solver Challenge (CSSC 2014; \citeR{hutter-aij17a}). In this case, the
	\runtime{}s on the training and test set were almost perfectly linearly correlated, with a
	Spearman correlation coefficient of $0.99$, i.e., the ranking of the
	configurations on both sets is nearly identical; this is also visualized in
	Figure~\ref{fig:overtuning_instances:rooks}. This is a very good case for
	applying algorithm configuration, and, correspondingly, in the CSSC 2014
	algorithm configuration yielded large improvements for this benchmark.
	\item[\lingeling{} on mixed SAT] We reconstructed a benchmark from
	\citeA{ansotegui-ijcai15a} in which they optimized
	\lingeling{}~\cite{biere-sat14a} on a mixed set of industrial SAT
	instances. Instead of randomly splitting the data into train and test
	instances, they first created a training set by removing
	hard instances (i.e., not solved within the cutoff time by reference solvers)
	and used these remaining hard instances as test instances. 
	Figure~\ref{fig:overtuning_instances:lingeling} shows that \smac{} improved
	the \runtime{} of \lingeling{} on the training set but that these
	improvements did not generalize to the test instances. In fact, the training and test scores of the optimized configurations (orange squares) 
	are only weakly correlated (Spearman correlation coefficient of $0.15$). 
	The benchmark's heterogeneity and the 
	%corresponding 
	mismatch between
	training and test set make this benchmark poorly suited for algorithm
	configuration.
	
	\item[\clasp{} on LABS] Figure~\ref{fig:overtuning_instances:labs} shows
	another benchmark from the CSSC: configuration of \clasp{} on SAT-encoded low
	autocorrelation binary sequence (LABS) benchmarks~\cite{mugrauer2013-2}.
	This illustrates a rare worst case for algorithm configuration, in which
	performance even degrades on the training set, which is possible due to
	\smac{}'s (and any other configurator's) racing approach: the configurator
	already changes the incumbent before all training instances have been
	evaluated, and if a subset is not representative of the full set this may lead
	to performance degradation on the full set.

	While we have occasionally observed such strong heterogeneity on instances with very
	heterogeneous sources, it was very surprising to observe this in a case
	where all instances stemmed from the same instance family.
	We therefore analyzed this benchmark further \cite{hutter-aij17a},
	showing that twice as many \smac{} runs with a fivefold larger configuration
	budget managed to improve training performance slightly. However, that
	improvement on the training set still did not generalize to the test set due to the
	benchmark's heterogeneity. (Although visually not apparent from Figure~\ref{fig:overtuning_instances:labs}, for this benchmark, the correlation between scores on training and test instances was quite low ($0.42$) for the $20\%$ best-performing randomly sampled	configurations).
	Again, for such heterogeneous benchmarks we
	recommend the usage of portfolio approaches.
\end{description}

\paragraph{Best Practice}
Over-tuning is often not easy to fully rule out by design,
since the effect can only be measured by assessing test performance after the configuration process completed (for example by scatter plots, such as in Figure~\ref{fig:overtuning_instances}).
Nevertheless, the following strategies minimize the risk of over-tuning (see also Section~\ref{sec:further}):

\begin{enumerate}
\denselist
  \item The training instances should be representative of the test instances;
  \item The training set should be relatively large (typically hundreds to
  thousands of instances) to increase the chance of being representative;
  \item The instance sets should stem from a similar application, use context,
  etc., increasing the likelihood that they have similar structures which can be
  exploited with similar solution strategies;
  \item If the instance set is heterogeneous, portfolio approaches~\cite{xu-jair08a,kadioglu-cp11a,malitsky-cp12a,lindauer-jair15a} or instance-specific algorithm configuration~\cite{xu-aaai10a,kadioglu-ecai10} should be used. 
\end{enumerate}

%%%%%%%%%%%%%%%%%%%%%%%%%%%%%%%%%%%%%%%%%%%%%%%%%%%%%%%%%%%%%%%%%%%%
\subsection{\emph{Pitfall 7: Over-tuning to a Particular Machine Type}}
\label{sub:over-tuning_machine}
%%%%%%%%%%%%%%%%%%%%%%%%%%%%%%%%%%%%%%%%%%%%%%%%%%%%%%%%%%%%%%%%%%%%

In the age of cloud computing and large compute clusters,
an obvious idea is to use these remotely-accessible compute resources
to benchmark algorithms and configure them. 
However, in the end, these remote machines are not always the production systems
the algorithms are used on in the end.
\citeA{geschwender-lion14a} indicated in a preliminary study that it
is possible in principle to configure algorithms in the cloud,
and that the found configurations perform well on another machine.
Unfortunately, recent other experiments showed that this does not hold for all kinds of algorithms --
for example, the performance of solvers for SAT~\cite{aigner-pos13a} 
and mixed integer programming~(\citeR{lodi-theory14a}; \shortciteR{koch-mpc11a}) can depend strongly 
on the used machine type (including hardware, operating system and 
installed software libraries). 

\paragraph{Effect}

Some algorithms are machine-dependent and obtain different results depending on the hardware they run on. Being unaware of this can ruin both, a successful application and a comparison of configuration methods, in two ways: Firstly, when configuring on one system the best found configuration might perform poorly on another system. Secondly, the ranking of the best found configurations of target algorithms on one system might change when rerunning the experiments on a different system.

\paragraph{Example}

An example for such machine-dependent algorithms are SAT solvers that are often highly optimized against cache misses~\cite{aigner-pos13a}.
To study the effect of different machines, we optimized three SAT solvers from
the configurable SAT solver challenge \cite{hutter-aij17a}, namely
\minisathack{} \cite{oh-sat14a}, \clasp{}~\cite{gebser-ai12} and
\lingeling~\cite{biere-sat14a} on \circuitfuzz{} instances~\cite{brummayer-sat10a}.
As different machine types, we used AWS m4.4xlarge instances with 2.4-GHz Intel
Xeon E5-2676 v3 CPUs with 30MB level-3 cache and the META-cluster at the
University of Freiburg with 2.6GHz Intel Xeon E5-2650v2 8-core CPUs with 20 MB L3 cache.
On both systems, we ran Ubuntu 14.04 64bit and allowed for a memory limit of $3$GB for each solver run.
The binaries were statically compiled such that they are not linked against different libraries on the different systems.
For each solver we ran $12$ independent \smac{} runs and validated the cost of the best found configuration for each solver on test instances on the same system.

\begin{table}[t]
\centering
\begin{tabular}{l|cc|cc}%cc}
    & \multicolumn{2}{c}{AWS} & \multicolumn{2}{c}{META-Cluster}\\ 
solver			& Rank & PAR10 & Rank & PAR10\\ 
\midrule
\minisathack    & 1 & 187 & 2 & 205\\
\clasp     		 & 2 & 215 & 1 & 193\\
\lingeling 		 & 3 & 231 & 3 & 208\\ 
\bottomrule
\end{tabular}
\caption{Three SAT solvers from the configurable SAT solver challenge on \circuitfuzz{} instances on two different hardware systems.}
\label{tab:over-tuning_machine}
\end{table}

Table~\ref{tab:over-tuning_machine} lists the ranking and the PAR10 scores of
the solvers on each machine (showing the test cost of the configuration
performing best on training); we note that the PAR10 scores are only
comparable on the same system.
In both environments, \lingeling{} ended up on rank 3, but the ranks of \clasp{} and \minisathack{} differed between the two environments: if the AWS cloud would be our environment for running AC experiments,
we would decide for \minisathack{},
but this would not be the best choice on the META-cluster. We note that, since we picked the best of 12 \smac{} runs, due to the high variance of extremal statistics, the exact numbers of this experiments might vary in a rerun. Since we did not have enough compute resources on AWS for carrying out multiple runs, to gain additional confidence in our conclusions, we carried out an additional experiment: we validated the configurations found on AWS on the META-cluster and found that in that setting the configured \minisathack{} performed even worse than \lingeling{} and \clasp{}. Therefore, we conclude that the ranking of configured algorithms depends on the hardware.
 
\paragraph{Best Practice}

We note that this pitfall exists only for %some 
machine-sensitive algorithms.
Therefore, we recommend to investigate whether an algorithm at hand has machine-dependent performance,
for example, by validating the performance of various configurations on both the
system used for configuration and the production system.

%%%%%%%%%%%%%%%%%%%%%%%%%%%%%%%%%%%%%%%%%%%%%%%%%%%%%%%%%%%%%%%%%%%%
\section{Further Recommendations for Effective Configuration}
\label{sec:further}
%%%%%%%%%%%%%%%%%%%%%%%%%%%%%%%%%%%%%%%%%%%%%%%%%%%%%%%%%%%%%%%%%%%%

In the following, we describe recommendations for users of algorithm
configuration systems to obtain parameter configurations that will perform
better in production.
Some of these recommendations are rules of thumb, since the involved factors for
a successful configuration can be very complex and can change across
configuration scenarios.
For general empirical algorithmics,
\citeA{mcgeoch-book12a} recommends further best practices, 
including design, reports and analysis of computational experiments.

\subsection{Training and Test Sets}\label{sec:train_test}
As discussed before, following standard practice, we strongly recommend to split the available instances into
a training and a test set to obtain an unbiased estimate of generalization
performance from the test set~\cite{birattari-tuning09a}.
To obtain trivial parallelization of randomized configuration procedures, we
recommend to run $n$ independent configuration runs and use the training set to
select the best of the $n$ resulting configurations~\cite{hutter-lion12a}. Only
that single chosen configuration should be evaluated on the test set; we explicitly note that we
cannot select the configuration that performs best on the test set, because
that would amount to peeking at our test data and render performance estimates
on the test set biased.

\subsection{Representative Instances and Runtime Cutoff}\label{sec:rep_insts_cutoff}
Intuitively, instances for which every parameter configuration times out do not help the configurator to make progress. One strategy can be to remove these from the training set. However, this comes with the risk to bias the training set towards easy instances and should be used with caution.
Generally, we therefore recommend to use training instances for the
configuration process that are representative of the ones to be solved later. Using training instances from a range
of hardness can also often help yield
configurations that generalize~\cite{hoos-lion13a}.
If feasible, we recommend to select instances and \runtime{} cutoffs such that roughly $75\%$ or more of the training instances used during configuration can be solved by the initial parameter configuration within the cutoff.
We emphasize that -- while the configuration protocol may in principle choose to
subsample the training instances in arbitrary ways -- the test set should never
be touched and not pre-evaluated to ensure an unbiased cost estimate of the optimized configurations in the end (see Pitfall~6).
To select a good training instance set, \citeA{bayless-lion14a} proposed a way to quantify whether an instance set is a good proxy for another instance set.
Furthermore, \citeA{styles-gecco13a} proposed a splitting strategy of the instances for better scaling to hard instances:
They split the instances into a training, validation and test set to
use easy instances during configuration for fast progress and select a
configuration on the harder validation set such that the configuration will
perform well on the hard test set.

\subsection{Homogeneous vs Heterogenous Instance Sets}\label{sec:homo}
Sometimes configurators are used to obtain well-performing and robust
configurations on a heterogeneous instance set. 
However, we know from algorithm selection~\cite{rice76a,kotthoff-aim14a}
that often no single configuration exists that performs well for
all instances in a heterogeneous set, but a portfolio of configurations
is required to obtain good performance~\cite{xu-rcra11a,kadioglu-ecai10}.
Furthermore, the task of algorithm configuration becomes a lot harder if all
instances can be solved best with very different configurations.
Therefore, we recommend to use algorithm configuration mainly on homogeneous instance sets.
Furthermore, the size of the used instance set should be adjusted accordingly to
the homogeneity of the instance set: on homogeneous instance sets, $50$
instances might suffice for good generalization performance to new instances,
but on fairly heterogeneous instance sets, we recommend to use at least $300$
or, if possible, more than $1000$ instances to obtain a robust parameter
configuration.

\subsection{Appropriate Configuration Settings}\label{sec:settings}
To use configurators,
the user has to set the budget available for the configurator.
If the configuration budget is too small, the configurator might make little or
no progress within it.
In contrast, if the configuration budget is too large, we waste a lot of time and computational resources
because the configurator might converge long before the budget is used up.
A good rule of thumb in our experience is to use a budget that equals at least
the expected \runtime{} of the default configuration on $200$ to $1000$
instances. 
In practice, an effective configuration budget strongly depends on several
factors, including heterogeneity of the instance set (more heterogeneous
instance sets require a larger configuration budget) or size of the
configuration space (larger configuration spaces require more time to
search effectively~\cite{hutter-aij17a}). Finally, if the configurator finds better performing configurations quickly, then the estimate of the total runtime based on the runtime of the default configuration might be too conservative.

\subsection{Efficient Use of Parallel Resources}\label{sec:parallel_res}
Some configurators (such as \gga{}, \irace{} and \dsmac{}) 
can make use of parallel resources,  while others (such as \paramils{}
and \smac{}) benefit from executing several independent parallel runs\footnote{In order to perform $k$ independent runs with
\paramils{} or \smac{}, one should use a different seed (equivalent to the
numRun parameter) for each run.} (and using
the result from the one with the best training set
performance; see, e.g., \citeA{hutter-lion12a}).  In the special case of
\gga{}, using more parallel resources can actually improve the adaptive capping mechanism. 
Given $k$ cores, we therefore recommend to execute one \gga{} run with $k$
cores, but $k$ independent \paramils{} or \smac{} runs with one core each.
While this protocol was not used in early
works\footnote{\citeA{hutter-lion11a} only used a single
core per run of \gga{}, but still followed the protocol by
\citeA{ansotegui-cp09a} to race groups of 8 runs in parallel per core;
therefore, \gga{}'s adaptive capping mechanism was the same in that work as in
\citeA{ansotegui-cp09a}.}, it has 
been used in more recent evaluations~\cite{ansotegui-ijcai15a,hutter-aij17a}.

\subsection{Reasonable Configuration Space}\label{sec:config_space}
Another challenge in using algorithm configuration systems is to find the
best configuration space. The user has to decide which parameters to optimize
and which ranges to allow. The optimal set of parameters to configure is often not clear and
in case of doubt, we recommend to add more parameters to the
configuration space and to use generous value ranges.

However, we note that unreasonably large configuration spaces are hard to
configure and require substantially larger configuration budgets.
For example, the state-of-the-art SAT solver \lingeling{}~\cite{biere-tech13a} has more than $300$ parameters
and most of them have a value range between $0$ and $32$bit maxint, but most of
these parameters are either not really relevant for optimizing \lingeling{}'s
\runtime{} or the relevant value ranges are much smaller.
Even though \lingeling{} can already substantially benefit from configuration we expect that with a more carefully designed configuration space even better results could be obtained. Therefore, we recommend to avoid including such parameters and to use smaller value ranges if corresponding expert knowledge is available.

Nevertheless, configurators have already been successfully applied to such large configuration spaces: \gga{}++ has been used to optimize over $100$ parameters of \lingeling{}~\cite{ansotegui-ijcai15a}, \irace{} has been used to optimize over $200$ parameters of the mixed integer programming solver SCIP~\cite{lopez-ejor14a,achterberg-mpc09a} and with \smac{}, we have optimized configuration spaces with over $900$ parameters~\cite{lindauer-aij17a}.

\subsection{Which Parameters to Tune}\label{sec:which_params}
Parameters should never be part of the configuration space 
if they change the semantics of the problem to be solved;
e.g., do not tune the allowed memory or parameters that control whether a
run is counted as successful (such as the allowed optimality gap in an
optimization setting).
Furthermore, to obtain an unbiased estimate of a configuration's performance
across seeds one should not include the seed (or parameters with a similar
effect) as a tunable parameter.

\subsection{Runtime Metrics}\label{sec:time_metric}
A common cost metric in algorithm configuration is \runtime{}. 
Obtaining clean \runtime{} measurements is a problem that is by no means limited to algorithm configuration and also appears in general empirical algorithmics~\cite{mcgeoch-book12a}.
However, in algorithm configuration, this problem can be even more tricky, 
because benchmark machines can be influenced by heavy I/O load on a shared file system created by multiple configuration runs (see Pitfall 3). Furthermore, other running processes on the same machine can influence the measurements. The latter issue can be fixed by using processor affinity to bind processes to a certain CPU.
Therefore, we recommend to measure CPU time instead of wallclock time. However, binding processes does not grant exclusive usage of the assigned cores; thus other interfering factors such as operation system load and shared caches remain.
Also, CPU time can sometimes be brittle; e.g., its resolution
can be insufficient for very short target algorithm runs, such as milliseconds.
We note that algorithm configuration \emph{can} be used to optimize runtime at such very small scales,
but extreme care needs to be taken to avoid any pitfalls associated to measuring runtimes.
When possible, a better solution for this case is to measure and optimize elementary operations, such as search steps of a local search algorithm or MEMS \cite<number of memory accesses,>{knuth-book11a}; however, it has to be ensured that such proxy
metrics correlate well with \runtime{}.
Additionally, expensive one-time operations, such as downloading files or setting up should not be part of the measured \runtime{} and need to be ignored, e.g. via the wrapper.
Finally, it remains an open question how robust are different ways to measure runtime and related metrics and how do they influence algorithm configuration.

\subsection{Monitoring Experiments} Even a well designed experiment can go wrong because of software and hardware issues. This makes conducting a flawless experiment challenging. However, the risk for falling for a pitfall can be minimized when carefully monitoring ongoing experiments. 

Investigating at the first bad sign can save a lot of time and resources. An unexpectedly high load on a machine or swapping memory can be signs of misconfigured scripts. More subtle effects that should also raise one's attention include the following: (1) the target algorithm uses much more wallclock time than the CPU time reported to the configurator; (2)~many configurations crash; or (3)~there is a large variation between the performances of independent configuration runs that only differ in their seeds.

We recommend to analyze ongoing experiments with respect to these signs and make use of automated tools, e.g. CAVE~\cite{biedenkapp-lion18}, to analyze and visualize experimental results in a common and unified way independently of the underlying configurator and problem.

\subsection{Comparing Configurators on Existing, Open-Source Benchmarks}\label{sec:aclib}
Although algorithm configuration has been well established for over a decade,
nearly every new paper on this topic uses a new set of benchmarks to compare different configurators.
This makes it harder to assess progress in the field, and every new benchmark
could again suffer from one of the pitfalls described above.
Therefore, we recommend to use existing and open-source algorithm configuration benchmarks
that are already well tested and can be freely used by the community.
The only existing library of such benchmarks we are aware of is the algorithm
configuration library AClib~\cite{hutter-lion14a}, which comprises $326$ benchmarks (in version 1.2) based on open-source scripts and allows users to pick benchmarks from different domains (e.g., mixed integer programming, AI Planning, SAT, and machine learning) and with different characteristics (e.g., small or large configuration spaces).

\section{A Generic Wrapper: Towards a Reliable and Unified AC Interface}
\label{sec:gen_wrapper}

Learning from the pitfalls above, our conclusion is that most of these pitfalls
can be either completely prevented or their risk of occurrence can be substantially reduced
by using a generic wrapper which wraps the executions of all target algorithm runs and
has the following features: 

\begin{enumerate}
\denselist
  \item Parsing the input arguments provided by the configurator in a uniform way 
        such that a user only needs to implement a function to translate them into a call of the target algorithm;
  \item Reliably limiting the run's computational resources (\runtime{} and memory consumption);
  \item Measuring the cost metric in a standardized way (for which a user only needs to implement a function to parse the output of the target algorithm); and
  \item Returning the output in a standardized way.
\end{enumerate}

\noindent We note that some pitfalls cannot be tested easily. E.g., the user is still responsible for domain-dependent solution checking and checking whether the configurator is used as intended. However, if using a wrapper with the features above most pitfalls can be avoided.
To demonstrate the usefulness of such a generic wrapper, and to provide a practical proposal for avoiding many of the described pitfalls, we implemented such a wrapper and are already using it in the algorithm configuration library AClib~\cite{hutter-lion14a}, to wrap $20$ different target algorithms.\footnote{Our package called \texttt{GenericWrapper4AC} is available at \url{https://github.com/automl/GenericWrapper4AC}.}
To address the pitfalls mentioned above, our generic wrapper implements the following best practices:

\begin{description}
	\item[Resource Limitation] The tool \runsolver{}~\cite{roussel-jsat11a} has been used for several years by the SAT community, in SAT competitions and by many SAT developers, to limit the runtime and memory consumption of an algorithm run.\footnote{The runsolver uses process group IDs to keep track of running processes For example, if the memory or time limit is exceeded, it traverses the process tree bottom-up to terminate all processes that run. However, we note that it is possible to bypass this procedure if a process forks itself or starts a process on a different machine, which can neither be detected nor monitored by the \runsolver{}.} We also use this tool in the generic wrapper to reliably limit such resources and to measure algorithm runtimes. This addresses both Pifall 1 (``Trusting Your Target Algorithm'') and Pitfall 2 (``Not Terminating Target Algorithm Runs Properly'').
	\item[Solution Checking for SAT] One of the exemplary instantiations of the generic wrapper we provide for SAT solvers implements solution checking to avoid issues of algorithm correctness (Pitfall 1: ``Trusting Your Target Algorithm'').
	\item[Writing to \$TMPDIR] On most high-performance clusters these days, the environment variable \$TMPDIR specifies a temporary directory on a local file system (not on a shared file system) of a compute node that allows for fast write and read access without affecting the remaining cluster. If this  environment variable is set, the generic wrapper writes all temporary files (e.g., log files of the \runsolver{}) to this folder. It only copies these files to a permanent file system in case of a crash of the target algorithm to allow debugging of these crashes. This addresses Pitfall 3 (``Slow File System'').
\end{description} 

\noindent Furthermore, the use of the generic wrapper has the following advantages compared to implementing the same features directly in an algorithm configurator (which is nevertheless a feasible approach for some use cases):

\begin{description}
	\item[Fair Comparisons] As discussed in Pitfall~4 (``Handling Target Algorithm Runs Differently''), to compare different configurators, using a uniform wrapper will ensure that all configurators optimize the same objective function. Even if a wrapper turns out to have a bug, at least all configurators would be affected in the same way.
	\item[Easy Use of Different Configurators] So far, most configurators implement different interfaces to call target algorithms. Therefore, users often implement only one of the interfaces and have not explored which of the available configurator is in fact the best one for their configuration problem. Using a generic wrapper (implementing either a unified interface or several configurator-specific interfaces) will also help users to easily use several configurators for their target algorithms. 
	\item[Easier Implementation of New Configurators] The implementation of new configurators is not an easy task, mainly because the handling of target algorithm runs may require many lines of code and is often still brittle. To reduce the burden on configurator developers, the generic wrapper can take over some of the functions required in this setting (e.g., resource limitations). Also, when translating a configurator to a new programming language, one can ensure that functionalities regarding handling that target algorithm remain exactly the same.
	\item[Open Source and Community] Since the generic wrapper is an open-source implementation, we believe that the community will improve the code base and thus improve its quality and robustness over time. 
\end{description} 

\noindent Appendix~\ref{appendix:details} provides additional details about our generic wrapper, and an example wrapper for a SAT solver.

\section{Conclusion}
Empirically comparing algorithms correctly is hard. This is well known and true for almost every empirical study that involves running third-party code, stochastic algorithms and computationally expensive computations and therefore also applies to algorithm configuration. 
Subtle mistakes, such as measuring the wrong metric or running parallel experiments without meticulous resource management, can heavily bias the outcome. In this work, we pointed out several pitfalls that can occur in running algorithm configuration experiments and provide concrete examples of how these can impact results. We found that many of these pitfalls result from treating the objective function differently in different configurators, from issues in allocating and monitoring resource consumption, and from various issues concerning over-tuning. To prevent most of these pitfalls we share recommendations and best practices for conducting algorithm configuration experiments, which we hope to be useful for both novices and experts. We also provide an open-source implementation of a generic wrapper that provides a unified interface for the communication between target algorithms and configurators and for limiting resource consumption.

\section*{Acknowledgements}

We thank Manuel L{\'{o}}pez{-}Ib{\'{a}}{\~{n}}ez and Kevin Tierney for 
adapting the interfaces of \irace{} and \gga{} to work together with
\texttt{GenericWrapper4AC}, Yuri Malitsky and Horst
Samulowitz for providing the wrappers and benchmarks of \citeA{ansotegui-ijcai15a},
and Kevin Tierney, Manuel L{\'{o}}pez{-}Ib{\'{a}}{\~{n}}ez and Lars
Kotthoff for very helpful feedback on the first draft of the paper that led
to the inclusion of some further possible issues. Some of the recommendations in
Section~\ref{sec:further} were inspired by a discussion at a 
Dagstuhl seminar (see~\citeA{lindauer-dagstuhl17a} for more details), and we are thankful for the valuable contributions of the attendees of that discussion:
Aymeric Blot, Wanru Gao, 
Holger Hoos, Laetitia Jourdan, Lars Kotthoff, Manuel L{\'o}pez-Ib{\'a}{\~n}ez, 
Nysret Musliu, G{\"u}nter Rudolph, Marc Schoenauer, Thomas St{\"u}tzle and Joaquin Vanschoren.
We also thank the anonymous reviewers for their valuable feedback. The authors acknowledge funding by the DFG (German Research Foundation) under
Emmy Noether grant HU 1900/2-1. K.\ Eggensperger additionally acknowledges
funding by the State Graduate Funding Program of Baden-Württemberg.

\appendix

\section{Details on \wrapper}
\label{appendix:details}

Listing~\ref{lst:wrapper:in} shows an example for how to extend the \wrapper{} 
to wrap the well-known SAT Solver \minisat~\cite{een-sat03}.
Since the output format is standardized in the SAT community,
we already provide a domain-specific generic wrapper, called
\texttt{SatWrapper}, which can parse and verify the SAT solver's output using
standard tools from the annual SAT competitions.
Therefore, SAT solver users only need to implement one method, which
constructs a command line call string for their SAT solver from the provided
input arguments (parameter settings, instance, cutoff time, seed).
\begin{lstlisting}[language=Python, caption={Example GenericWrapper for SAT Solver
\minisat{}, building on our domain-specific \texttt{SatWrapper}}, label=lst:wrapper:in]
class MiniSATWrapper(SatWrapper):

 	def get_command_line_args(self, runargs, config):
		cmd = "minisat -rnd-seed=%d" %(runargs["seed"])       
		for name, value in config.items():
			cmd += " %s=%s" %(name,  value)
		cmd += " %s" %(runargs["instance"])
		return cmd
\end{lstlisting}
In the example shown, the command line call of \minisat{} consists of passing
the random \emph{seed} (Line 4), adding all parameters in the format
\texttt{parameter=value} (Lines 5 and 6), and appending the CNF instance name at
the end (Line 7). 
Importantly, it takes care of all aspects of handling cutoff times, measuring \runtime{}s, etc, to avoid the pitfalls discussed in Section~\ref{sec:pitfalls_wrapper}.
\begin{lstlisting}[language=Python, caption={Example GenericWrapper from scratch}, label=lst:wrapper:out] 
class SimpleWrapper(AbstractWrapper):
    
	def get_command_line_args(self, runargs, config):
		[...]
		
	def process_results(self, fp, exit_code):
		try:
		    resultMap = {'status': 'SUCCESS', 'cost': float(fp.read()) }
		except ValueError:
		    resultMap = {'status': 'CRASHED'}
		
		return resultMap
\end{lstlisting}
For users of algorithm configuration outside SAT solving,
Listing~\ref{lst:wrapper:out} shows an example for how to write a function
\texttt{process\_results} to parse algorithm outputs.
Let us assume that the target algorithm only prints the target cost to be
minimized (similar to the format of \irace{}~\cite{lopez-ibanez-orp16}).
Reading the output of the provided file pointer \texttt{fp},
the function builds and returns a dictionary
which includes the cost value and a status, which is either \texttt{SUCCESS} if the target algorithm printed
only a single number or \texttt{CRASHED} otherwise.
Other states can be \texttt{TIMEOUT} for using more than the
cutoff time $\cutoff$ or \texttt{ABORT} to signal the configurator to abort the AC experiment because of major issues.
Furthermore, the exit code of the target algorithm run is also provided (but not
used in our example).
Another possible functionality that is not shown here is to implement a (domain-specific) method
to verify the target algorithm's returned solution.

Except these two target algorithm-specific functions,
the \wrapper{} handles everything else, including
\begin{itemize}
  \item Parsing the input format; native interfaces to \paramils{}, \roar{} and \smac{} are supported right now, and an additional layer to run \gga{}(++) and \irace{} is available as well.
  (see AClib2\footnote{\url{https://bitbucket.org/mlindauer/aclib2}} for examples).
  \item Calling the target algorithm and limiting its resource limits using the
  \runsolver{} tool~\cite{roussel-jsat11a}
  \item Measuring the CPU time of the target algorithm run (using \runsolver)
  \item Returning the cost of the target algorithm run to the configurator
\end{itemize} 

The \wrapper{} is available at GitHub
and can be easily installed via \texttt{python setup.py install} (including the \runsolver{}) and runs on UNIX systems. 

\bibliography{clean}

\begin{thebibliography}{}

\bibitem[\protect\BCAY{Achterberg}{Achterberg}{2009}]{achterberg-mpc09a}
Achterberg, T. \BBOP2009\BBCP.
\newblock \BBOQ {SCIP}: solving constraint integer programs\BBCQ\
\newblock {\Bem Mathematical Programming Computation}, {\Bem 1}, 1--41.

\bibitem[\protect\BCAY{Aigner, Biere, Kirsch, Niemetz,\ \BBA\ Preiner}{Aigner
  et~al.}{2013}]{aigner-pos13a}
Aigner, M., Biere, A., Kirsch, C., Niemetz, A., \BBA\ Preiner, M.
  \BBOP2013\BBCP.
\newblock \BBOQ Analysis of portfolio-style parallel {SAT} solving on current
  multi-core architectures\BBCQ\
\newblock In {\Bem Proceeding of the Fourth International Workshop on
  Pragmatics of SAT (POS'13)}.

\bibitem[\protect\BCAY{Ansel, Kamil, Veeramachaneni, Ragan{-}Kelley, Bosboom,
  O'Reilly,\ \BBA\ Amarasinghe}{Ansel et~al.}{2014}]{ansel-pact14a}
Ansel, J., Kamil, S., Veeramachaneni, K., Ragan{-}Kelley, J., Bosboom, J.,
  O'Reilly, U., \BBA\ Amarasinghe, S. \BBOP2014\BBCP.
\newblock \BBOQ Opentuner: an extensible framework for program autotuning\BBCQ\
\newblock In Amaral, J.\BBACOMMA\  \BBA\ Torrellas, J.\BEDS, {\Bem Proceedings
  of the International Conference on Parallel Architectures and Compilation
  (PACT)}, \BPGS\ 303--316. {ACM}.

\bibitem[\protect\BCAY{Ans{\'{o}}tegui, Gab{\`{a}}s, Malitsky,\ \BBA\
  Sellmann}{Ans{\'{o}}tegui et~al.}{2016}]{ansotegui-aij16}
Ans{\'{o}}tegui, C., Gab{\`{a}}s, J., Malitsky, Y., \BBA\ Sellmann, M.
  \BBOP2016\BBCP.
\newblock \BBOQ Maxsat by improved instance-specific algorithm
  configuration\BBCQ\
\newblock {\Bem Artificial Intelligence}, {\Bem 235}, 26--39.

\bibitem[\protect\BCAY{Ans{\'o}tegui, Malitsky, Sellmann,\ \BBA\
  Tierney}{Ans{\'o}tegui et~al.}{2015}]{ansotegui-ijcai15a}
Ans{\'o}tegui, C., Malitsky, Y., Sellmann, M., \BBA\ Tierney, K.
  \BBOP2015\BBCP.
\newblock \BBOQ Model-based genetic algorithms for algorithm
  configuration\BBCQ\
\newblock In Yang, Q.\BBACOMMA\  \BBA\ Wooldridge, M.\BEDS, {\Bem Proceedings
  of the 25th International Joint Conference on Artificial Intelligence
  (IJCAI'15)}, \BPGS\ 733--739.

\bibitem[\protect\BCAY{Ans{\'o}tegui, Sellmann,\ \BBA\ Tierney}{Ans{\'o}tegui
  et~al.}{2009}]{ansotegui-cp09a}
Ans{\'o}tegui, C., Sellmann, M., \BBA\ Tierney, K. \BBOP2009\BBCP.
\newblock \BBOQ A gender-based genetic algorithm for the automatic
  configuration of algorithms\BBCQ\
\newblock In Gent, I.\BED, {\Bem Proceedings of the Fifteenth International
  Conference on Principles and Practice of Constraint Programming (CP'09)},
  \lowercase{\BVOL}\ 5732 of {\Bem Lecture Notes in Computer Science}, \BPGS\
  142--157. Springer-Verlag.

\bibitem[\protect\BCAY{Audemard\ \BBA\ Simon}{Audemard\ \BBA\
  Simon}{2009}]{audemard-ijcai09a}
Audemard, G.\BBACOMMA\  \BBA\ Simon, L. \BBOP2009\BBCP.
\newblock \BBOQ Predicting learnt clauses quality in modern {SAT} solvers\BBCQ\
\newblock In Boutilier, C.\BED, {\Bem Proceedings of the 22th International
  Joint Conference on Artificial Intelligence (IJCAI'09)}, \BPGS\ 399--404.

\bibitem[\protect\BCAY{Bartz-Beielstein, Lasarczyk,\ \BBA\
  Preuss}{Bartz-Beielstein et~al.}{2010}]{bartzbeielstein-emaoa10a}
Bartz-Beielstein, T., Lasarczyk, C., \BBA\ Preuss, M. \BBOP2010\BBCP.
\newblock \BBOQ The sequential parameter optimization toolbox\BBCQ\
\newblock In Bartz-Beielstein, T., Chiarandini, M., Paquete, L., \BBA\ Preus,
  M.\BEDS, {\Bem Experimental Methods for the Analysis of Optimization
  Algorithms}, \BPGS\ 337--362. Springer-Verlag.

\bibitem[\protect\BCAY{Bayless, Tompkins,\ \BBA\ Hoos}{Bayless
  et~al.}{2014}]{bayless-lion14a}
Bayless, S., Tompkins, D., \BBA\ Hoos, H. \BBOP2014\BBCP.
\newblock \BBOQ Evaluating instance generators by configuration\BBCQ\
\newblock In Pardalos, P.\BBACOMMA\  \BBA\ Resende, M.\BEDS, {\Bem Proceedings
  of the Eighth International Conference on Learning and Intelligent
  Optimization (LION'14)}, Lecture Notes in Computer Science. Springer-Verlag.

\bibitem[\protect\BCAY{Bezerra, L{\'{o}}pez{-}Ib{\'{a}}{\~{n}}ez,\ \BBA\
  St{\"{u}}tzle}{Bezerra et~al.}{2016}]{bezerra-tec16a}
Bezerra, L., L{\'{o}}pez{-}Ib{\'{a}}{\~{n}}ez, M., \BBA\ St{\"{u}}tzle, T.
  \BBOP2016\BBCP.
\newblock \BBOQ Automatic component-wise design of multiobjective evolutionary
  algorithms\BBCQ\
\newblock {\Bem {IEEE} Trans. Evolutionary Computation}, {\Bem 20\/}(3),
  403--417.

\bibitem[\protect\BCAY{Biedenkapp, Marben, Lindauer,\ \BBA\ Hutter}{Biedenkapp
  et~al.}{2018}]{biedenkapp-lion18}
Biedenkapp, A., Marben, J., Lindauer, M., \BBA\ Hutter, F. \BBOP2018\BBCP.
\newblock \BBOQ Cave: Configuration assessment, visualization and
  evaluation\BBCQ\
\newblock In {\Bem Proceedings of the Tenth International Conference on
  Learning and Intelligent Optimization (LION'18)}, Lecture Notes in Computer
  Science. Springer-Verlag.
\newblock To appear.

\bibitem[\protect\BCAY{Biere}{Biere}{2013}]{biere-tech13a}
Biere, A. \BBOP2013\BBCP.
\newblock \BBOQ {Lingeling}, {Plingeling} and {Treengeling} entering the {SAT}
  competition 2013\BBCQ\
\newblock In Balint, A., Belov, A., Heule, M., \BBA\ J{\"a}rvisalo, M.\BEDS,
  {\Bem Proceedings of {SAT} Competition 2013: Solver and Benchmark
  Descriptions}, \lowercase{\BVOL}\ B-2013-1 of {\Bem Department of Computer
  Science Series of Publications B}, \BPGS\ 51--52. University of Helsinki.

\bibitem[\protect\BCAY{Biere}{Biere}{2014}]{biere-sat14a}
Biere, A. \BBOP2014\BBCP.
\newblock \BBOQ Yet another local search solver and {Lingeling} and friends
  entering the {SAT} competition 2014\BBCQ\
\newblock In Belov, A., Diepold, D., Heule, M., \BBA\ J{\"a}rvisalo, M.\BEDS,
  {\Bem Proceedings of {SAT} Competition 2014: Solver and Benchmark
  Descriptions}, \lowercase{\BVOL}\ B-2014-2 of {\Bem Department of Computer
  Science Series of Publications B}, \BPGS\ 39--40. University of Helsinki.

\bibitem[\protect\BCAY{Biere, Heule, {van Maaren},\ \BBA\ Walsh}{Biere
  et~al.}{2009}]{biere-satbook}
Biere, A., Heule, M., {van Maaren}, H., \BBA\ Walsh, T.\BEDS. \BBOP2009\BBCP.
\newblock {\Bem Handbook of Satisfiability}, \lowercase{\BVOL}\ 185 of {\Bem
  Frontiers in Artificial Intelligence and Applications}.
\newblock IOS Press.

\bibitem[\protect\BCAY{Birattari\ \BBA\ Kacprzyk}{Birattari\ \BBA\
  Kacprzyk}{2009}]{birattari-tuning09a}
Birattari, M.\BBACOMMA\  \BBA\ Kacprzyk, J. \BBOP2009\BBCP.
\newblock {\Bem Tuning metaheuristics: a machine learning perspective},
  \lowercase{\BVOL}\ 197.
\newblock Springer-Verlag.

\bibitem[\protect\BCAY{Birattari, Stützle, Paquete,\ \BBA\
  Varrentrapp}{Birattari et~al.}{2002}]{birattari-gecco02a}
Birattari, M., Stützle, T., Paquete, L., \BBA\ Varrentrapp, K. \BBOP2002\BBCP.
\newblock \BBOQ A racing algorithm for configuring metaheuristics\BBCQ\
\newblock In Langdon, W., Cantu-Paz, E., Mathias, K., Roy, R., Davis, D., Poli,
  R., Balakrishnan, K., Honavar, V., Rudolph, G., Wegener, J., Bull, L.,
  Potter, M., Schultz, A., Miller, J., Burke, E., \BBA\ Jonoska, N.\BEDS, {\Bem
  Proceedings of the Genetic and Evolutionary Computation Conference
  (GECCO'02)}, \BPGS\ 11--18. Morgan Kaufmann Publishers.

\bibitem[\protect\BCAY{Birattari}{Birattari}{2004}]{birattari-PhD2004}
Birattari, M. \BBOP2004\BBCP.
\newblock {\Bem The problem of tuning metaheuristics as seen from a machine
  learning perspective}.
\newblock Ph.D.\ thesis, Universit\'e Libre de Bruxelles.

\bibitem[\protect\BCAY{Brochu, Cora,\ \BBA\ de~Freitas}{Brochu
  et~al.}{2010}]{brochu-arXiv10a}
Brochu, E., Cora, V., \BBA\ de~Freitas, N. \BBOP2010\BBCP.
\newblock \BBOQ A tutorial on {Bayesian} optimization of expensive cost
  functions, with application to active user modeling and hierarchical
  reinforcement learning\BBCQ.
\newblock {\Bem arXiv:1012.2599}.

\bibitem[\protect\BCAY{Brummayer, Lonsing,\ \BBA\ Biere}{Brummayer
  et~al.}{2012}]{brummayer-sat10a}
Brummayer, R., Lonsing, F., \BBA\ Biere, A. \BBOP2012\BBCP.
\newblock \BBOQ Automated testing and debugging of {SAT} and {QBF}
  solvers\BBCQ\
\newblock In Cimatti, A.\BBACOMMA\  \BBA\ Sebastiani, R.\BEDS, {\Bem
  Proceedings of the Fifteenth International Conference on Theory and
  Applications of Satisfiability Testing (SAT'12)}, \lowercase{\BVOL}\ 7317 of
  {\Bem Lecture Notes in Computer Science}, \BPGS\ 44--57. Springer-Verlag.

\bibitem[\protect\BCAY{C{\'{a}}ceres\ \BBA\ St{\"{u}}tzle}{C{\'{a}}ceres\ \BBA\
  St{\"{u}}tzle}{2017}]{caceres-endm17a}
C{\'{a}}ceres, L.~P.\BBACOMMA\  \BBA\ St{\"{u}}tzle, T. \BBOP2017\BBCP.
\newblock \BBOQ Exploring variable neighborhood search for automatic algorithm
  configuration\BBCQ\
\newblock {\Bem Electronic Notes in Discrete Mathematics}, {\Bem 58}, 167--174.

\bibitem[\protect\BCAY{Chiarandini, Fawcett,\ \BBA\ Hoos}{Chiarandini
  et~al.}{2008}]{chiarandini-patat08a}
Chiarandini, M., Fawcett, C., \BBA\ Hoos, H. \BBOP2008\BBCP.
\newblock \BBOQ A modular multiphase heuristic solver for post enrolment course
  timetabling\BBCQ\
\newblock In Gendreau, M.\BBACOMMA\  \BBA\ Burke, E.\BEDS, {\Bem Proceedings of
  the Seventh International Conference on the Practice and Theory of Automated
  Timetabling}.

\bibitem[\protect\BCAY{Dorigo}{Dorigo}{2016}]{dorigo-sw16}
Dorigo, M. \BBOP2016\BBCP.
\newblock \BBOQ Swarm intelligence: A few things you need to know if you want
  to publish in this journal\BBCQ.

\bibitem[\protect\BCAY{E{\'{e}}n\ \BBA\ S{\"{o}}rensson}{E{\'{e}}n\ \BBA\
  S{\"{o}}rensson}{2004}]{een-sat03}
E{\'{e}}n, N.\BBACOMMA\  \BBA\ S{\"{o}}rensson, N. \BBOP2004\BBCP.
\newblock \BBOQ An extensible {SAT}-solver\BBCQ\
\newblock In Giunchiglia, E.\BBACOMMA\  \BBA\ Tacchella, A.\BEDS, {\Bem
  Proceedings of the conference on Theory and Applications of Satisfiability
  Testing (SAT)}, \lowercase{\BVOL}\ 2919 of {\Bem Lecture Notes in Computer
  Science}, \BPGS\ 502--518. Springer-Verlag.

\bibitem[\protect\BCAY{Falkner, Lindauer,\ \BBA\ Hutter}{Falkner
  et~al.}{2015}]{falkner-sat15a}
Falkner, S., Lindauer, M., \BBA\ Hutter, F. \BBOP2015\BBCP.
\newblock \BBOQ {SpySMAC}: Automated configuration and performance analysis of
  {SAT} solvers\BBCQ\
\newblock In Heule, M.\BBACOMMA\  \BBA\ Weaver, S.\BEDS, {\Bem Proceedings of
  the Eighteenth International Conference on Theory and Applications of
  Satisfiability Testing (SAT'15)}, Lecture Notes in Computer Science, \BPGS\
  1--8. Springer-Verlag.

\bibitem[\protect\BCAY{Feurer, Springenberg,\ \BBA\ Hutter}{Feurer
  et~al.}{2015}]{feurer-aaai15a}
Feurer, M., Springenberg, T., \BBA\ Hutter, F. \BBOP2015\BBCP.
\newblock \BBOQ Initializing {B}ayesian hyperparameter optimization via
  meta-learning\BBCQ\
\newblock In Bonet, B.\BBACOMMA\  \BBA\ Koenig, S.\BEDS, {\Bem Proceedings of
  the Twenty-nineth National Conference on Artificial Intelligence (AAAI'15)},
  \BPGS\ 1128--1135. AAAI Press.

\bibitem[\protect\BCAY{Gebser, Kaminski, Kaufmann, Schaub, Schneider,\ \BBA\
  Ziller}{Gebser et~al.}{2011}]{gebser-lpnmr11a}
Gebser, M., Kaminski, R., Kaufmann, B., Schaub, T., Schneider, M., \BBA\
  Ziller, S. \BBOP2011\BBCP.
\newblock \BBOQ A portfolio solver for answer set programming: Preliminary
  report\BBCQ\
\newblock In Delgrande, J.\BBACOMMA\  \BBA\ Faber, W.\BEDS, {\Bem Proceedings
  of the Eleventh International Conference on Logic Programming and
  Nonmonotonic Reasoning (LPNMR'11)}, \lowercase{\BVOL}\ 6645 of {\Bem Lecture
  Notes in Computer Science}, \BPGS\ 352--357. Springer-Verlag.

\bibitem[\protect\BCAY{Gebser, Kaufmann,\ \BBA\ Schaub}{Gebser
  et~al.}{2012}]{gebser-ai12}
Gebser, M., Kaufmann, B., \BBA\ Schaub, T. \BBOP2012\BBCP.
\newblock \BBOQ Conflict-driven answer set solving: From theory to
  practice\BBCQ\
\newblock {\Bem Artificial Intelligence}, {\Bem 187-188}, 52--89.

\bibitem[\protect\BCAY{Gent, Grant, MacIntyre, Prosser, Shaw, Smith,\ \BBA\
  Walsh}{Gent et~al.}{1997}]{gent-report97a}
Gent, I., Grant, S., MacIntyre, E., Prosser, P., Shaw, P., Smith, B., \BBA\
  Walsh, T. \BBOP1997\BBCP.
\newblock \BBOQ How not to do it\BBCQ\
\newblock \BTR\ 97.92, University of Leeds.

\bibitem[\protect\BCAY{Geschwender, Hutter, Kotthoff, Malitsky, Hoos,\ \BBA\
  Leyton{-}Brown}{Geschwender et~al.}{2014}]{geschwender-lion14a}
Geschwender, D., Hutter, F., Kotthoff, L., Malitsky, Y., Hoos, H., \BBA\
  Leyton{-}Brown, K. \BBOP2014\BBCP.
\newblock \BBOQ Algorithm configuration in the cloud: A feasibility study\BBCQ\
\newblock In Pardalos, P.\BBACOMMA\  \BBA\ Resende, M.\BEDS, {\Bem Proceedings
  of the Eighth International Conference on Learning and Intelligent
  Optimization (LION'14)}, Lecture Notes in Computer Science, \BPGS\ 41--46.
  Springer-Verlag.

\bibitem[\protect\BCAY{Hansen}{Hansen}{2006}]{hansen-eda06}
Hansen, N. \BBOP2006\BBCP.
\newblock \BBOQ The {CMA} evolution strategy: a comparing review\BBCQ\
\newblock In Lozano, J., Larranaga, P., Inza, I., \BBA\ Bengoetxea, E.\BEDS,
  {\Bem Towards a new evolutionary computation. Advances on estimation of
  distribution algorithms}, \BPGS\ 75--102. Springer.

\bibitem[\protect\BCAY{Henderson, Islam, Bachman, Pineau, Precup,\ \BBA\
  Meger}{Henderson et~al.}{2018}]{henderson-arxiv13a}
Henderson, P., Islam, R., Bachman, P., Pineau, J., Precup, D., \BBA\ Meger, D.
  \BBOP2018\BBCP.
\newblock \BBOQ Deep reinforcement learning that matters\BBCQ.
\newblock {\Bem arXiv:1709.06560}.

\bibitem[\protect\BCAY{Heule, Hunt,\ \BBA\ Wetzler}{Heule
  et~al.}{2014}]{heule-stvr14a}
Heule, M., Hunt, W., \BBA\ Wetzler, N. \BBOP2014\BBCP.
\newblock \BBOQ Bridging the gap between easy generation and efficient
  verification of unsatisfiability proofs\BBCQ\
\newblock {\Bem Software Testing Verification and Reliability}, {\Bem 24\/}(8),
  593--607.

\bibitem[\protect\BCAY{Hooker}{Hooker}{1995}]{hooker-jh95a}
Hooker, J. \BBOP1995\BBCP.
\newblock \BBOQ Testing heuristics: We have it all wrong\BBCQ\
\newblock {\Bem Journal of Heuristics}, {\Bem 1}, 33--42.

\bibitem[\protect\BCAY{Hoos}{Hoos}{2012}]{hoos-cacm12}
Hoos, H. \BBOP2012\BBCP.
\newblock \BBOQ Programming by optimization\BBCQ\
\newblock {\Bem Communications of the ACM}, {\Bem 55\/}(2), 70--80.

\bibitem[\protect\BCAY{Hoos, Kaufmann, Schaub,\ \BBA\ Schneider}{Hoos
  et~al.}{2013}]{hoos-lion13a}
Hoos, H., Kaufmann, B., Schaub, T., \BBA\ Schneider, M. \BBOP2013\BBCP.
\newblock \BBOQ Robust benchmark set selection for boolean constraint
  solvers\BBCQ\
\newblock In Pardalos, P.\BBACOMMA\  \BBA\ Nicosia, G.\BEDS, {\Bem Proceedings
  of the Seventh International Conference on Learning and Intelligent
  Optimization (LION'13)}, \lowercase{\BVOL}\ 7997 of {\Bem Lecture Notes in
  Computer Science}, \BPGS\ 138--152. Springer-Verlag.

\bibitem[\protect\BCAY{Howe\ \BBA\ Dahlman}{Howe\ \BBA\
  Dahlman}{2002}]{howe-jair02a}
Howe, A.\BBACOMMA\  \BBA\ Dahlman, E. \BBOP2002\BBCP.
\newblock \BBOQ A critical assessment of benchmark comparison in planning\BBCQ\
\newblock {\Bem Journal of Artificial Intelligence Research}, {\Bem 17}, 1--33.

\bibitem[\protect\BCAY{Hutter, Babi\'c, Hoos,\ \BBA\ Hu}{Hutter
  et~al.}{2007a}]{hutter-fmcad07a}
Hutter, F., Babi\'c, D., Hoos, H., \BBA\ Hu, A. \BBOP2007a\BBCP.
\newblock \BBOQ Boosting verification by automatic tuning of decision
  procedures\BBCQ\
\newblock In O'Conner, L.\BED, {\Bem Formal Methods in Computer Aided Design
  (FMCAD'07)}, \BPGS\ 27--34. IEEE Computer Society Press.

\bibitem[\protect\BCAY{Hutter, Hoos,\ \BBA\ Leyton-Brown}{Hutter
  et~al.}{2010a}]{hutter-cpaior10a}
Hutter, F., Hoos, H., \BBA\ Leyton-Brown, K. \BBOP2010a\BBCP.
\newblock \BBOQ Automated configuration of mixed integer programming
  solvers\BBCQ\
\newblock In Lodi, A., Milano, M., \BBA\ Toth, P.\BEDS, {\Bem Proceedings of
  the Seventh International Conference on Integration of {AI} and {OR}
  Techniques in Constraint Programming (CPAIOR'10)}, \lowercase{\BVOL}\ 6140 of
  {\Bem Lecture Notes in Computer Science}, \BPGS\ 186--202. Springer-Verlag.

\bibitem[\protect\BCAY{Hutter, Hoos,\ \BBA\ Leyton-Brown}{Hutter
  et~al.}{2010b}]{hutter-amai10}
Hutter, F., Hoos, H., \BBA\ Leyton-Brown, K. \BBOP2010b\BBCP.
\newblock \BBOQ Tradeoffs in the empirical evaluation of competing algorithm
  designs\BBCQ\
\newblock {\Bem Annals of Mathematics and Artificial Intelligenc (AMAI),
  Special Issue on Learning and Intelligent Optimization}, {\Bem 60\/}(1),
  65--89.

\bibitem[\protect\BCAY{Hutter, Hoos,\ \BBA\ Leyton-Brown}{Hutter
  et~al.}{2011}]{hutter-lion11a}
Hutter, F., Hoos, H., \BBA\ Leyton-Brown, K. \BBOP2011\BBCP.
\newblock \BBOQ Sequential model-based optimization for general algorithm
  configuration\BBCQ\
\newblock In Coello, C.\BED, {\Bem Proceedings of the Fifth International
  Conference on Learning and Intelligent Optimization (LION'11)},
  \lowercase{\BVOL}\ 6683 of {\Bem Lecture Notes in Computer Science}, \BPGS\
  507--523. Springer-Verlag.

\bibitem[\protect\BCAY{Hutter, Hoos,\ \BBA\ Leyton-Brown}{Hutter
  et~al.}{2012}]{hutter-lion12a}
Hutter, F., Hoos, H., \BBA\ Leyton-Brown, K. \BBOP2012\BBCP.
\newblock \BBOQ Parallel algorithm configuration\BBCQ\
\newblock In Hamadi, Y.\BBACOMMA\  \BBA\ Schoenauer, M.\BEDS, {\Bem Proceedings
  of the Sixth International Conference on Learning and Intelligent
  Optimization (LION'12)}, \lowercase{\BVOL}\ 7219 of {\Bem Lecture Notes in
  Computer Science}, \BPGS\ 55--70. Springer-Verlag.

\bibitem[\protect\BCAY{Hutter, Hoos, Leyton-Brown,\ \BBA\ St{\"u}tzle}{Hutter
  et~al.}{2009}]{hutter-jair09a}
Hutter, F., Hoos, H., Leyton-Brown, K., \BBA\ St{\"u}tzle, T. \BBOP2009\BBCP.
\newblock \BBOQ Param{ILS}: An automatic algorithm configuration
  framework\BBCQ\
\newblock {\Bem Journal of Artificial Intelligence Research}, {\Bem 36},
  267--306.

\bibitem[\protect\BCAY{Hutter, Hoos,\ \BBA\ St\"utzle}{Hutter
  et~al.}{2007b}]{hutter-aaai07a}
Hutter, F., Hoos, H., \BBA\ St\"utzle, T. \BBOP2007b\BBCP.
\newblock \BBOQ Automatic algorithm configuration based on local search\BBCQ\
\newblock In Holte, R.\BBACOMMA\  \BBA\ Howe, A.\BEDS, {\Bem Proceedings of the
  Twenty-second National Conference on Artificial Intelligence (AAAI'07)},
  \BPGS\ 1152--1157. AAAI Press.

\bibitem[\protect\BCAY{Hutter, Lindauer, Balint, Bayless, Hoos,\ \BBA\
  Leyton-Brown}{Hutter et~al.}{2017}]{hutter-aij17a}
Hutter, F., Lindauer, M., Balint, A., Bayless, S., Hoos, H., \BBA\
  Leyton-Brown, K. \BBOP2017\BBCP.
\newblock \BBOQ The configurable {SAT} solver challenge ({CSSC})\BBCQ\
\newblock {\Bem Artificial Intelligence}, {\Bem 243}, 1--25.

\bibitem[\protect\BCAY{Hutter, L\'{o}pez-Ib\'{a}nez, Fawcett, Lindauer, Hoos,
  Leyton-Brown,\ \BBA\ St\"utzle}{Hutter et~al.}{2014a}]{hutter-lion14a}
Hutter, F., L\'{o}pez-Ib\'{a}nez, M., Fawcett, C., Lindauer, M., Hoos, H.,
  Leyton-Brown, K., \BBA\ St\"utzle, T. \BBOP2014a\BBCP.
\newblock \BBOQ {AClib}: a benchmark library for algorithm configuration\BBCQ\
\newblock In Pardalos, P.\BBACOMMA\  \BBA\ Resende, M.\BEDS, {\Bem Proceedings
  of the Eighth International Conference on Learning and Intelligent
  Optimization (LION'14)}, Lecture Notes in Computer Science. Springer-Verlag.

\bibitem[\protect\BCAY{Hutter, Tompkins,\ \BBA\ Hoos}{Hutter
  et~al.}{2002}]{hutter-cp02a}
Hutter, F., Tompkins, D., \BBA\ Hoos, H. \BBOP2002\BBCP.
\newblock \BBOQ Scaling and probabilistic smoothing: Efficient dynamic local
  search for {SAT}\BBCQ\
\newblock In Hentenryck, P.~V.\BED, {\Bem Proceedings of the international
  conference on Principles and Practice of Constraint Programming},
  \lowercase{\BVOL}\ 2470 of {\Bem Lecture Notes in Computer Science}, \BPGS\
  233--248. Springer-Verlag.

\bibitem[\protect\BCAY{Hutter, Xu, Hoos,\ \BBA\ Leyton-Brown}{Hutter
  et~al.}{2014b}]{hutter-aij14a}
Hutter, F., Xu, L., Hoos, H., \BBA\ Leyton-Brown, K. \BBOP2014b\BBCP.
\newblock \BBOQ Algorithm runtime prediction: Methods and evaluation\BBCQ\
\newblock {\Bem Artificial Intelligence}, {\Bem 206}, 79--111.

\bibitem[\protect\BCAY{Kadioglu, Malitsky, Sabharwal, Samulowitz,\ \BBA\
  Sellmann}{Kadioglu et~al.}{2011}]{kadioglu-cp11a}
Kadioglu, S., Malitsky, Y., Sabharwal, A., Samulowitz, H., \BBA\ Sellmann, M.
  \BBOP2011\BBCP.
\newblock \BBOQ Algorithm selection and scheduling\BBCQ\
\newblock In Lee, J.\BED, {\Bem Proceedings of the Seventeenth International
  Conference on Principles and Practice of Constraint Programming (CP'11)},
  \lowercase{\BVOL}\ 6876 of {\Bem Lecture Notes in Computer Science}, \BPGS\
  454--469. Springer-Verlag.

\bibitem[\protect\BCAY{Kadioglu, Malitsky, Sellmann,\ \BBA\ Tierney}{Kadioglu
  et~al.}{2010}]{kadioglu-ecai10}
Kadioglu, S., Malitsky, Y., Sellmann, M., \BBA\ Tierney, K. \BBOP2010\BBCP.
\newblock \BBOQ {ISAC} - instance-specific algorithm configuration\BBCQ\
\newblock In Coelho, H., Studer, R., \BBA\ Wooldridge, M.\BEDS, {\Bem
  Proceedings of the Nineteenth European Conference on Artificial Intelligence
  (ECAI'10)}, \BPGS\ 751--756. IOS Press.

\bibitem[\protect\BCAY{Knuth}{Knuth}{2011}]{knuth-book11a}
Knuth, D. \BBOP2011\BBCP.
\newblock {\Bem The Art of Computer Programming, Volume {IV}}.
\newblock Addison-Wesley.

\bibitem[\protect\BCAY{Koch, Achterberg, Andersen, Bastert, Berthold, Bixby,
  Danna, Gamrath, Gleixner, Heinz, Lodi, Mittelmann, Ralphs, Salvagnin,
  Steffy,\ \BBA\ Wolter}{Koch et~al.}{2011}]{koch-mpc11a}
Koch, T., Achterberg, T., Andersen, E., Bastert, O., Berthold, T., Bixby, R.,
  Danna, E., Gamrath, G., Gleixner, A., Heinz, S., Lodi, A., Mittelmann, H.,
  Ralphs, T., Salvagnin, D., Steffy, D., \BBA\ Wolter, K. \BBOP2011\BBCP.
\newblock \BBOQ {MIPLIB} 2010\BBCQ\
\newblock {\Bem Mathematical Programming Computation}, {\Bem 3}, 103--163.

\bibitem[\protect\BCAY{Kotthoff}{Kotthoff}{2014}]{kotthoff-aim14a}
Kotthoff, L. \BBOP2014\BBCP.
\newblock \BBOQ Algorithm selection for combinatorial search problems: A
  survey\BBCQ\
\newblock {\Bem AI Magazine}, {\Bem 35\/}(3), 48--60.

\bibitem[\protect\BCAY{Laguna}{Laguna}{2017}]{laguna-jh}
Laguna, M. \BBOP2017\BBCP.
\newblock \BBOQ Journal of heuristic policies on heuristic search
  research\BBCQ.

\bibitem[\protect\BCAY{Lindauer, Hoos, Hutter,\ \BBA\ Schaub}{Lindauer
  et~al.}{2015}]{lindauer-jair15a}
Lindauer, M., Hoos, H., Hutter, F., \BBA\ Schaub, T. \BBOP2015\BBCP.
\newblock \BBOQ Autofolio: An automatically configured algorithm selector\BBCQ\
\newblock {\Bem Journal of Artificial Intelligence Research}, {\Bem 53},
  745--778.

\bibitem[\protect\BCAY{Lindauer, Hoos, Leyton-Brown,\ \BBA\ Schaub}{Lindauer
  et~al.}{2017a}]{lindauer-aij17a}
Lindauer, M., Hoos, H., Leyton-Brown, K., \BBA\ Schaub, T. \BBOP2017a\BBCP.
\newblock \BBOQ Automatic construction of parallel portfolios via algorithm
  configuration\BBCQ\
\newblock {\Bem Artificial Intelligence}, {\Bem 244}, 272--290.

\bibitem[\protect\BCAY{Lindauer\ \BBA\ Hutter}{Lindauer\ \BBA\
  Hutter}{2017b}]{lindauer-dagstuhl17a}
Lindauer, M.\BBACOMMA\  \BBA\ Hutter, F. \BBOP2017b\BBCP.
\newblock \BBOQ Pitfalls and best practices for algorithm configuration
  (breakout session report)\BBCQ\
\newblock {\Bem Dagstuhl Reports}, {\Bem 6}, 70--72.

\bibitem[\protect\BCAY{Lodi\ \BBA\ Tramontani}{Lodi\ \BBA\
  Tramontani}{2014}]{lodi-theory14a}
Lodi, A.\BBACOMMA\  \BBA\ Tramontani, A. \BBOP2014\BBCP.
\newblock \BBOQ Performance variability in mixed-integer programming\BBCQ\
\newblock In Topaloglu, H., Smith, J., \BBA\ Greenberg, H.\BEDS, {\Bem Theory
  Driven by Influential Applications}, \BCH~1, \BPGS\ 1--12. INFORMS.

\bibitem[\protect\BCAY{L{\'{o}}pez{-}Ib{\'{a}}{\~{n}}ez, Dubois-Lacoste,
  Caceres, Birattari,\ \BBA\ St{\"{u}}tzle}{L{\'{o}}pez{-}Ib{\'{a}}{\~{n}}ez
  et~al.}{2016}]{lopez-ibanez-orp16}
L{\'{o}}pez{-}Ib{\'{a}}{\~{n}}ez, M., Dubois-Lacoste, J., Caceres, L.~P.,
  Birattari, M., \BBA\ St{\"{u}}tzle, T. \BBOP2016\BBCP.
\newblock \BBOQ The irace package: Iterated racing for automatic algorithm
  configuration\BBCQ\
\newblock {\Bem Operations Research Perspectives}, {\Bem 3}, 43--58.

\bibitem[\protect\BCAY{López-Ibáñez\ \BBA\ Stützle}{López-Ibáñez\ \BBA\
  Stützle}{2014}]{lopez-ejor14a}
López-Ibáñez, M.\BBACOMMA\  \BBA\ Stützle, T. \BBOP2014\BBCP.
\newblock \BBOQ Automatically improving the anytime behaviour of optimisation
  algorithms\BBCQ\
\newblock {\Bem European Journal of Operational Research}, {\Bem 235},
  569--582.

\bibitem[\protect\BCAY{Malitsky, Sabharwal, Samulowitz,\ \BBA\
  Sellmann}{Malitsky et~al.}{2012}]{malitsky-cp12a}
Malitsky, Y., Sabharwal, A., Samulowitz, H., \BBA\ Sellmann, M. \BBOP2012\BBCP.
\newblock \BBOQ Parallel {SAT} solver selection and scheduling\BBCQ\
\newblock In Milano, M.\BED, {\Bem Proceedings of the Eighteenth International
  Conference on Principles and Practice of Constraint Programming (CP'12)},
  \lowercase{\BVOL}\ 7514 of {\Bem Lecture Notes in Computer Science}, \BPGS\
  512--526. Springer-Verlag.

\bibitem[\protect\BCAY{Manthey}{Manthey}{2014b}]{riss}
Manthey, N. \BBOP2014b\BBCP.
\newblock \BBOQ {R}iss 4.27\BBCQ\
\newblock In Belov, A., Diepold, D., Heule, M., \BBA\ J{\"a}rvisalo, M.\BEDS,
  {\Bem Proceedings of {SAT} Competition 2014: Solver and Benchmark
  Descriptions}, \lowercase{\BVOL}\ B-2014-2 of {\Bem Department of Computer
  Science Series of Publications B}, \BPGS\ 65--67. University of Helsinki.

\bibitem[\protect\BCAY{Manthey\ \BBA\ Lindauer}{Manthey\ \BBA\
  Lindauer}{2016}]{manthey-sat16a}
Manthey, N.\BBACOMMA\  \BBA\ Lindauer, M. \BBOP2016\BBCP.
\newblock \BBOQ Spybug: Automated bug detection in the configuration space of
  sat solvers\BBCQ\
\newblock In Creignou, N.\BBACOMMA\  \BBA\ Berre, D.~L.\BEDS, {\Bem Proceedings
  of the Nineteenth International Conference on Theory and Applications of
  Satisfiability Testing (SAT'16)}, Lecture Notes in Computer Science, \BPGS\
  554--561. Springer-Verlag.

\bibitem[\protect\BCAY{Manthey\ \BBA\ Steinke}{Manthey\ \BBA\
  Steinke}{2014a}]{manthey-sat14r}
Manthey, N.\BBACOMMA\  \BBA\ Steinke, P. \BBOP2014a\BBCP.
\newblock \BBOQ Too many rooks\BBCQ\
\newblock In Belov, A., Diepold, D., Heule, M., \BBA\ J{\"a}rvisalo, M.\BEDS,
  {\Bem Proceedings of {SAT} Competition 2014: Solver and Benchmark
  Descriptions}, \lowercase{\BVOL}\ B-2014-2 of {\Bem Department of Computer
  Science Series of Publications B}, \BPGS\ 97--98. University of Helsinki.

\bibitem[\protect\BCAY{McGeoch}{McGeoch}{1987}]{mcgeoch-phd87}
McGeoch, C. \BBOP1987\BBCP.
\newblock {\Bem Experimental Analysis of Algorithms}.
\newblock Ph.D.\ thesis, Carnegie-Mellon University, Computer Science.

\bibitem[\protect\BCAY{McGeoch}{McGeoch}{2002}]{mcggeoch-hgo02a}
McGeoch, C. \BBOP2002\BBCP.
\newblock \BBOQ Experimental analysis of algorithms\BBCQ\
\newblock In Pardalos, P.\BBACOMMA\  \BBA\ Romeijn, E.\BEDS, {\Bem Handbook of
  Global Optimization}, \BPGS\ 489--513. Springer-Verlag.

\bibitem[\protect\BCAY{McGeoch}{McGeoch}{2012}]{mcgeoch-book12a}
McGeoch, C.~C. \BBOP2012\BBCP.
\newblock {\Bem A Guide to Experimental Algorithmics}.
\newblock Cambridge University Press.

\bibitem[\protect\BCAY{Mockus, Tiesis,\ \BBA\ Zilinskas}{Mockus
  et~al.}{1978}]{mockus-tgo78a}
Mockus, J., Tiesis, V., \BBA\ Zilinskas, A. \BBOP1978\BBCP.
\newblock \BBOQ The application of {B}ayesian methods for seeking the
  extremum\BBCQ\
\newblock {\Bem Towards Global Optimization}, {\Bem 2\/}(117-129).

\bibitem[\protect\BCAY{Mugrauer\ \BBA\ Balint}{Mugrauer\ \BBA\
  Balint}{2013}]{mugrauer2013-2}
Mugrauer, F.\BBACOMMA\  \BBA\ Balint, A. \BBOP2013\BBCP.
\newblock \BBOQ {SAT} encoded low autocorrelation binary sequence ({LABS})
  benchmark description\BBCQ\
\newblock In Balint, A., Belov, A., Heule, M., \BBA\ J{\"a}rvisalo, M.\BEDS,
  {\Bem Proceedings of {SAT} Competition 2013: Solver and Benchmark
  Descriptions}, \lowercase{\BVOL}\ B-2013-1 of {\Bem Department of Computer
  Science Series of Publications B}. University of Helsinki.

\bibitem[\protect\BCAY{Oh}{Oh}{2014}]{oh-sat14a}
Oh, C. \BBOP2014\BBCP.
\newblock \BBOQ {MiniSat HACK 999ED, MiniSat HACK 1430ED and SWDiA5BY}\BBCQ\
\newblock In Belov, A., Diepold, D., Heule, M., \BBA\ J{\"a}rvisalo, M.\BEDS,
  {\Bem Proceedings of {SAT} Competition 2014: Solver and Benchmark
  Descriptions}, \lowercase{\BVOL}\ B-2014-2 of {\Bem Department of Computer
  Science Series of Publications B}, \BPG~46. University of Helsinki.

\bibitem[\protect\BCAY{Rice}{Rice}{1976}]{rice76a}
Rice, J. \BBOP1976\BBCP.
\newblock \BBOQ The algorithm selection problem\BBCQ\
\newblock {\Bem Advances in Computers}, {\Bem 15}, 65--118.

\bibitem[\protect\BCAY{Roussel}{Roussel}{2011}]{roussel-jsat11a}
Roussel, O. \BBOP2011\BBCP.
\newblock \BBOQ Controlling a solver execution with the runsolver tool\BBCQ\
\newblock {\Bem Journal on Satisfiability, Boolean Modeling and Computation},
  {\Bem 7\/}(4), 139--144.

\bibitem[\protect\BCAY{Schneider\ \BBA\ Hoos}{Schneider\ \BBA\
  Hoos}{2012}]{hoos-lion12a}
Schneider, M.\BBACOMMA\  \BBA\ Hoos, H. \BBOP2012\BBCP.
\newblock \BBOQ Quantifying homogeneity of instance sets for algorithm
  configuration\BBCQ\
\newblock In Hamadi, Y.\BBACOMMA\  \BBA\ Schoenauer, M.\BEDS, {\Bem Proceedings
  of the Sixth International Conference on Learning and Intelligent
  Optimization (LION'12)}, \lowercase{\BVOL}\ 7219 of {\Bem Lecture Notes in
  Computer Science}, \BPGS\ 190--204. Springer-Verlag.

\bibitem[\protect\BCAY{Shahriari, Swersky, Wang, Adams,\ \BBA\
  de~Freitas}{Shahriari et~al.}{2016}]{shahriari-ieee16a}
Shahriari, B., Swersky, K., Wang, Z., Adams, R., \BBA\ de~Freitas, N.
  \BBOP2016\BBCP.
\newblock \BBOQ Taking the human out of the loop: {A} review of {B}ayesian
  optimization\BBCQ\
\newblock {\Bem Proceedings of the {IEEE}}, {\Bem 104\/}(1), 148--175.

\bibitem[\protect\BCAY{Snoek, Larochelle,\ \BBA\ Adams}{Snoek
  et~al.}{2012}]{snoek-nips12a}
Snoek, J., Larochelle, H., \BBA\ Adams, R.~P. \BBOP2012\BBCP.
\newblock \BBOQ Practical {B}ayesian optimization of machine learning
  algorithms\BBCQ\
\newblock In Bartlett, P., Pereira, F., Burges, C., Bottou, L., \BBA\
  Weinberger, K.\BEDS, {\Bem Proceedings of the 26th International Conference
  on Advances in Neural Information Processing Systems (NIPS'12)}, \BPGS\
  2960--2968.

\bibitem[\protect\BCAY{Soos}{Soos}{2014}]{cryptominisat}
Soos, M. \BBOP2014\BBCP.
\newblock \BBOQ {C}rypto{M}ini{S}at v4\BBCQ\
\newblock In Belov, A., Diepold, D., Heule, M., \BBA\ J{\"a}rvisalo, M.\BEDS,
  {\Bem Proceedings of {SAT} Competition 2014: Solver and Benchmark
  Descriptions}, \lowercase{\BVOL}\ B-2014-2 of {\Bem Department of Computer
  Science Series of Publications B}, \BPG~23. University of Helsinki.

\bibitem[\protect\BCAY{Styles\ \BBA\ Hoos}{Styles\ \BBA\
  Hoos}{2015}]{styles-gecco13a}
Styles, J.\BBACOMMA\  \BBA\ Hoos, H. \BBOP2015\BBCP.
\newblock \BBOQ Ordered racing protocols for automatically configuring
  algorithms for scaling performance\BBCQ\
\newblock In Blum, C.\BBACOMMA\  \BBA\ Alba, E.\BEDS, {\Bem Proceedings of the
  Genetic and Evolutionary Computation Conference (GECCO'13)}, \BPGS\ 551--558.
  {ACM}.

\bibitem[\protect\BCAY{Thornton, Hutter, Hoos,\ \BBA\ Leyton-Brown}{Thornton
  et~al.}{2013}]{thornton-kdd13a}
Thornton, C., Hutter, F., Hoos, H., \BBA\ Leyton-Brown, K. \BBOP2013\BBCP.
\newblock \BBOQ {A}uto-{WEKA}: combined selection and hyperparameter
  optimization of classification algorithms\BBCQ\
\newblock In Dhillon, I., Koren, Y., Ghani, R., Senator, T., Bradley, P.,
  Parekh, R., He, J., Grossman, R., \BBA\ Uthurusamy, R.\BEDS, {\Bem The 19th
  ACM SIGKDD International Conference on Knowledge Discovery and Data Mining
  (KDD'13)}, \BPGS\ 847--855. ACM Press.

\bibitem[\protect\BCAY{Tompkins\ \BBA\ Hoos}{Tompkins\ \BBA\
  Hoos}{2005}]{tompkins-sat05a}
Tompkins, D.\BBACOMMA\  \BBA\ Hoos, H. \BBOP2005\BBCP.
\newblock \BBOQ {UBCSAT}: An implementation and experimentation environment for
  {SLS} algorithms for {SAT} and {MAX-SAT}\BBCQ\
\newblock In {\Bem Proceedings of the Seventh International Conference on
  Theory and Applications of Satisfiability Testing ({SAT 2004})}, Lecture
  Notes in Computer Science, \BPGS\ 306--320. Springer-Verlag.

\bibitem[\protect\BCAY{Vallati, Fawcett, Gerevini, Hoos,\ \BBA\ Saetti}{Vallati
  et~al.}{2013}]{vallati-socs13a}
Vallati, M., Fawcett, C., Gerevini, A., Hoos, H., \BBA\ Saetti, A.
  \BBOP2013\BBCP.
\newblock \BBOQ Automatic generation of efficient domain-optimized planners
  from generic parametrized planners\BBCQ\
\newblock In Helmert, M.\BBACOMMA\  \BBA\ R{\"{o}}ger, G.\BEDS, {\Bem
  Proceedings of the Sixth Annual Symposium on Combinatorial Search (SOCS'14)}.
  {AAAI} Press.

\bibitem[\protect\BCAY{Weise, Chiong,\ \BBA\ Tang}{Weise
  et~al.}{2012}]{weise-jcst12a}
Weise, T., Chiong, R., \BBA\ Tang, K. \BBOP2012\BBCP.
\newblock \BBOQ Evolutionary optimization: Pitfalls and booby traps\BBCQ\
\newblock {\Bem Journal of Computer Science and Technology}, {\Bem 27},
  907--936.

\bibitem[\protect\BCAY{Xu, Hoos,\ \BBA\ Leyton-Brown}{Xu
  et~al.}{2010}]{xu-aaai10a}
Xu, L., Hoos, H., \BBA\ Leyton-Brown, K. \BBOP2010\BBCP.
\newblock \BBOQ Hydra: Automatically configuring algorithms for portfolio-based
  selection\BBCQ\
\newblock In Fox, M.\BBACOMMA\  \BBA\ Poole, D.\BEDS, {\Bem Proceedings of the
  Twenty-fourth National Conference on Artificial Intelligence (AAAI'10)},
  \BPGS\ 210--216. AAAI Press.

\bibitem[\protect\BCAY{Xu, Hutter, Hoos,\ \BBA\ Leyton-Brown}{Xu
  et~al.}{2008}]{xu-jair08a}
Xu, L., Hutter, F., Hoos, H., \BBA\ Leyton-Brown, K. \BBOP2008\BBCP.
\newblock \BBOQ {SAT}zilla: Portfolio-based algorithm selection for {SAT}\BBCQ\
\newblock {\Bem Journal of Artificial Intelligence Research}, {\Bem 32},
  565--606.

\bibitem[\protect\BCAY{Xu, Hutter, Hoos,\ \BBA\ Leyton-Brown}{Xu
  et~al.}{2011}]{xu-rcra11a}
Xu, L., Hutter, F., Hoos, H., \BBA\ Leyton-Brown, K. \BBOP2011\BBCP.
\newblock \BBOQ {Hydra-MIP}: Automated algorithm configuration and selection
  for mixed integer programming\BBCQ\
\newblock In {\Bem Proc. of RCRA workshop at IJCAI}.

\end{thebibliography}
%\bibliography{clean}
\bibliographystyle{theapa}

\end{document}